\documentclass{article} 
\usepackage{iclr2027_conference,times}

\usepackage{amsmath,amsfonts,bm}

\def\eqref#1{equation~\ref{#1}}

\def\1{\bm{1}}

\DeclareMathAlphabet{\mathsfit}{\encodingdefault}{\sfdefault}{m}{sl}
\SetMathAlphabet{\mathsfit}{bold}{\encodingdefault}{\sfdefault}{bx}{n}

\usepackage{hyperref}
\usepackage{url}

\usepackage{graphicx}
\usepackage{amsmath}
\usepackage{amssymb}
\usepackage{booktabs}
\usepackage{enumitem}
\usepackage{multirow}
\usepackage{wrapfig}
\usepackage{makecell}
\usepackage[accsupp]{axessibility}
\usepackage{footnote}
\usepackage{comment}
\usepackage{xcolor}
\usepackage{soul, color}

\usepackage{algorithm}
\usepackage{algorithmicx}
\usepackage{algpseudocode}
\usepackage{arydshln} 
\usepackage[
  separate-uncertainty = true,
  multi-part-units = repeat
]{siunitx}

\usepackage[capitalize]{cleveref}
\crefname{section}{Sec.}{Secs.}
\Crefname{section}{Section}{Sections}
\Crefname{table}{Table}{Tables}
\crefname{table}{Tab.}{Tabs.}

\usepackage{xspace}

\newcommand{\nickname}{EmbodiExp\xspace}

\title{\nickname: Compositional Experience Transfer for Embodied Agents}

\author{Zhifeng Gu \textsuperscript{1,2} \quad Ying Chen \textsuperscript{2} \quad Rui Jiang \textsuperscript{2} \quad Lei Yi \textsuperscript{2} \quad Bing Wang \textsuperscript{1}\thanks{Corresponding Author} \\
\textsuperscript{1} Spatial Intelligence Group, The Hong Kong Polytechnic University \\
\textsuperscript{2} XPENG Robotics \\
\texttt{zhifeng.gu@connect.polyu.hk \quad bingwang@polyu.edu.hk}
}

\iclrfinalcopy 
\begin{document}

\maketitle
\lhead{Preprint} 

\begin{abstract}

Experience from past tasks can help embodied agents perform new tasks, but it may not remain applicable or effective when conditions change.
Tasks with different overall goals often share subgoals, providing a basis for cross-task experience transfer. 
We present \textbf{\nickname}, a framework for constructing \emph{outcome-grounded subgoal experiences} and composing them in new tasks.
Each experience describes how to pursue a subgoal, when the guidance applies, and what outcome must be verified for completion.
Verified outcomes from applying one experience can satisfy the conditions for applying another, allowing experiences from different past tasks to be composed online.
By examining the actions and observed outcomes at the subgoal level, \nickname retains experience from successful subgoals and uses explicit action failures to identify actions to avoid under specific conditions.
Experiments on EmbodiedBench and VIGIL assess cross-task transfer with experiences kept fixed throughout held-out evaluation.
\nickname achieves success rates of 68.0\% on EB-Habitat and 68.3\% on EB-ALFRED, compared with 50.0\% and 54.7\% for the agent without stored experience.


\end{abstract}

\section{Introduction}
\label{sec:introduction}



Recent foundation models \citep{qwen36, openai2026gpt54} have advanced agentic robotics by equipping embodied agents with strong reasoning and high-level planning capabilities \citep{team2025gemini, yang2025embodiedbench}.
However, knowing \emph{what to do} does not imply knowing \emph{how to realize it} under the current environmental conditions \citep{ahn2022saycan, agarwal2025l3m+}.
To bridge this gap, embodied agents can leverage \emph{experience} accumulated through past interactions as empirical action-outcome knowledge \citep{wang2023voyager, shinn2023reflexion, park2023generative}.
At the core of experience is its \emph{transferability}: whether it can effectively inform grounded execution in new tasks.


To support such transferability, recent work has explored different perspectives, including retrieving task-relevant trajectories \citep{wang2026trajrag, yue2025mllm}, abstracting past executions into compact experiences \citep{ren2026worldmind, wang2026museagent}, and extracting reusable skills \citep{jiang2026xskill, ju2026embodiskill}.
Despite these advances, transferability remains a central challenge: experience must retain knowledge that extends beyond its source interaction, while its application must remain grounded in the conditions and outcomes of the new task.
This raises a fundamental question: \emph{what should be retained as experience, and how should it be applied?}


Experience can transfer across tasks through shared local goals, even when their overall goals and execution sequences differ \citep{konidaris2007building, james2020learning}.
However, sharing a local goal alone is insufficient for reliable transfer: a strategy effective in one interaction may not apply under different conditions, and even an applicable strategy may fail to produce the intended outcome \citep{ahn2022saycan, huang2023innermonologue}.
Transferable experience should thus specify not only how to pursue a goal, but also when the strategy applies and what outcome establishes goal completion \citep{konidaris2018skills}.
These requirements make the \emph{subgoal} level a natural granularity for experience, as each subgoal specifies a coherent intermediate goal whose completion can be assessed from local interaction evidence, independently of the source task's overall success or failure \citep{li2024embodied, hu2025hiagent}.
We therefore represent experience at the subgoal level, grounding its application and completion in explicit conditions and outcomes.


In this paper, we propose \textbf{\nickname}, a framework for constructing transferable experiences and composing them online during embodied task execution.
At its core is \emph{outcome-grounded subgoal experience}, which represents each experience through a subgoal, applicability conditions, an adaptable policy, and an intended local outcome.
\nickname constructs such experiences from local action-outcome evidence and retrieves them according to the current subgoal during execution.
An experience is applied only when its conditions are satisfied, and subgoal completion is established only when newly acquired interaction evidence supports the intended outcome.
This design enables two key mechanisms: experience composition across trajectories and failure utilization through selective retention of useful knowledge from failed trajectories.

\begin{figure*}[t]
\centering 
\centerline{\includegraphics[width=1\textwidth]{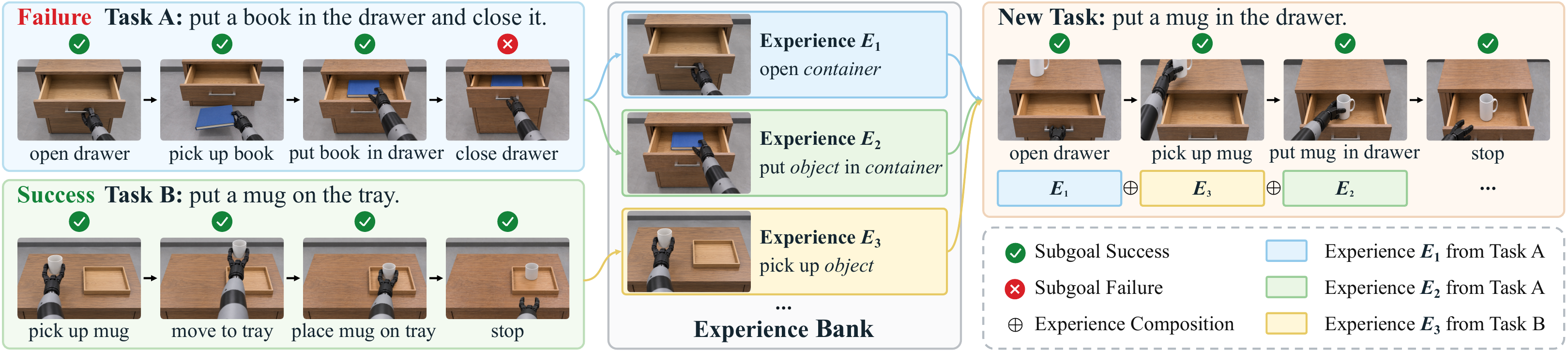}}
\vspace{-0.3cm}
\caption{
\textbf{Motivating Example: Compositional Experience Transfer across Tasks.} 
Different tasks can share subgoals even when their overall goals and execution sequences differ. 
The new task is successfully completed by composing experiences from different previous tasks.
}
\label{fig:teaser}
\vspace{-0.3cm}
\end{figure*}

For \textbf{experience composition}, a new task may involve subgoals whose relevant experiences originate from different source trajectories (Figure~\ref{fig:teaser}).
Retrieving relevant experience is insufficient, because retrieval alone does not determine when each experience becomes applicable as execution unfolds.
\nickname therefore \emph{composes} experience online through verified outcomes: the outcome realized by one experience may satisfy the applicability conditions of another, making the latter applicable as execution progresses.
This connects experiences across trajectories through grounded state transitions rather than through static combinations based on semantic
similarity alone.


For \textbf{failure utilization}, task-level failure provides only a coarse outcome signal, since a failed trajectory may contain both completed subgoals and failed local attempts.
Discarding the entire trajectory would lose useful local knowledge, whereas indiscriminate transfer could propagate behavior that did not achieve its intended outcome.
\nickname therefore performs subgoal-level outcome attribution on both successful and failed trajectories, admitting experience according to local action-outcome evidence rather than overall task outcomes.
Within a shared experience representation, verified local successes provide guidance on how to achieve a subgoal, while explicit failure evidence provides condition-specific guidance on operations to avoid during future execution.

Overall, our contributions are summarized as follows:
\begin{itemize}[leftmargin=*]
\setlength{\itemsep}{1pt}
\setlength{\parsep}{1pt}
\setlength{\parskip}{1pt}

    \item We present \textbf{\nickname}, a framework for constructing \emph{outcome-grounded subgoal experiences} from past task executions and transferring them to new tasks.
    
    \item We develop online experience composition across different trajectories, where verified local outcomes from one experience can establish the applicability conditions of another.
    
    \item We propose subgoal-level outcome attribution for failure utilization, retaining locally successful behavior and deriving condition-specific avoidance guidance from explicit failure evidence.
    
\end{itemize}

Extensive experimental evaluations on multiple benchmarks demonstrate the effectiveness of \nickname under a \emph{fold-based experience evaluation protocol} with frozen experience banks.

\section{Related Work}
\label{sec:literature}

\textbf{Multimodal Embodied Agents.}
Foundation models enable embodied agents to perform task planning and affordance-grounded decision making \citep{ahn2022saycan, song2023llmplanner}, generate executable programs \citep{liang2023codeaspolicies, singh2023progprompt}, and adapt execution using environment feedback \citep{huang2023innermonologue, wang2023deps}.
Vision-language models support embodied reasoning \citep{driess2023palme, chen2026capek}, while generalist robot policies map multimodal inputs to control \citep{brohan2023rt2, zitkovich2023rtx, kim2024openvla, octo2024}.
More recently, robot-use agents operate robots through perception and control interfaces.
Benchmarks assess household interaction, navigation, task completion, self-termination, and active spatial reasoning \citep{shridhar2020alfred, shridhar2021alfworld, yang2025embodiedbench, chen2026done, hong2026esi}.
Our work complements these systems through the construction and cross-task application of transferable experience from grounded interactions.

\textbf{Agent Memory.}
Agent memory retains past interactions as episodic records, summaries, and structured knowledge \citep{park2023generative, packer2023memgpt, zhong2024memorybank, chhikara2025mem0, xu2025amem, zhang2025gmemory}, with extensions for multimodal consolidation \citep{liu2025memverse} and learnable memory operations \citep{zhang2026memskill}.
In embodied environments, memory provides spatiotemporal context \citep{yang2025threedmem, hu2025threedllmmem, wang2026lmee, wang2024karma}, interaction histories \citep{torne2026mem}, hierarchical experience abstraction \citep{ma2026brainmem}, and interaction-derived object knowledge \citep{banwasi2026messymem}.
Within this broader literature, we focus on extracting outcome-grounded subgoal experiences from past interactions and composing them according to the current task state to guide execution across tasks.

\textbf{Experience and Skill Learning.}
Experience-driven agents distill past interactions into reflections and procedural guidance \citep{shinn2023reflexion, zhao2024expel, tang2025agentkb, ouyang2025reasoningbank, cao2025reme, cai2025flex, zhang2026liveevo} or build workflows and skills \citep{wang2024awm, fang2026memp, wang2023voyager, ni2026trace2skill}.
Further work explores feedback-driven skill refinement \citep{xia2026skillrl, zhang2026coevoskills, yang2026autoskill, zhang2026skill1}, evidence-backed skills with applicability and verification rules \citep{tang2026msce}, hierarchical skills \citep{sun2026trace2tower}, and how skill granularity and representation affect transfer \citep{feng2026break}.
MuSEAgent \citep{wang2026museagent} retrieves state-conditioned experiences, while XSkill \citep{jiang2026xskill} combines action-level experiences with task-level skills.
In embodied settings, WorldMind \citep{ren2026worldmind} distills process and goal experience, ELITE \citep{wei2026elite} retrieves intent-aligned strategies, MineEvolve \citep{xie2026mineevolve} builds skills and remedies from subgoal feedback, and EmbodiSkill \citep{ju2026embodiskill} revises skills through targeted error attribution.
\nickname constructs \emph{outcome-grounded subgoal experience} through local outcome attribution and composes it across tasks using applicability checks and outcome verification.

\section{Method}
\label{sec:method}

\subsection{Problem Formulation}
\label{sec:problem}

Consider an embodied agent $\mathcal{A}$ that receives a task instruction $\mathcal{I}$ and can generate high-level plans while interacting with an environment.
Such plans specify what to achieve, but grounded execution additionally requires selecting actions suited to the current environmental conditions and verifying that they produce the intended outcomes.
We refer to this gap between semantic task planning and grounded execution as the \emph{grounded-execution gap}.
Our objective is to bridge this gap by extracting transferable experience from prior interactions and applying it to new tasks.

\textbf{Interaction Trajectory.}
Let $o_t$ denote the agent's observation, including any available environment feedback, and $a_t$ its native action at interaction step $t$.
We represent an interaction trajectory as
\begin{equation}
    \tau =
    \left(
        \mathcal{I},
        o_0, a_0, o_1, \ldots, a_{T-1}, o_T,
        y_\tau
    \right),
    \label{eq:trajectory}
\end{equation}
where $T$ is the episode length and $y_\tau \in \{0,1\}$ indicates task-level success.
This terminal label does not specify the outcomes of individual subgoal attempts.

\textbf{Embodied Experience.}
A trajectory records a particular execution.
We define \emph{embodied experience} as empirical action-outcome knowledge accumulated from prior interactions to inform future grounded execution.
To support transfer across tasks, we seek a representation that connects execution guidance to its applicability conditions and a verifiable completion outcome.

\textbf{Outcome-Grounded Subgoal Experience.}
A \emph{subgoal} is an intermediate goal in a task plan, characterized by an intended local outcome and pursued through one or more native actions.
A \emph{subgoal attempt} is a contiguous interaction span directed toward a single subgoal.
We instantiate embodied experience at the subgoal level and represent each experience as
\begin{equation}
    e_i =
    \left(
        g_i, R_i, \pi_i, P_i
    \right),
    \label{eq:subgoal_experience}
\end{equation}
corresponding to \textsc{Subgoal}, \textsc{Requires}, \textsc{Policy}, and \textsc{Produces}, respectively.
Here, $g_i$ specifies the subgoal; 
$R_i$ describes the conditions for applying the experience; 
$\pi_i$ provides execution guidance on effective behavior or condition-specific operations to avoid;
and $P_i$ specifies the local outcome that must be verified for completion.
Together, these fields characterize \emph{what} to achieve, \emph{when} the experience applies, \emph{how} to pursue the subgoal, and \emph{what outcome} establishes its completion.

For an execution span directed toward $g_i$ from step $t$ to $t'$, let $\Omega_{t:t'}^{\mathrm{pub}}$ denote the agent-accessible interaction evidence collected during this span.
We define the outcome-verification indicator as
\begin{equation}
    c_i^{t:t'} =
    W_{\mathrm{pub}}
    \left(
        P_i, \Omega_{t:t'}^{\mathrm{pub}}
    \right)
    \in \{0,1\},
    \label{eq:experience_completion}
\end{equation}
where $W_{\mathrm{pub}}$ checks whether the evidence supports $P_i$.
A value of $1$ indicates verified completion, whereas $0$ indicates that completion remains unverified, not necessarily that the attempt has failed.

An experience is \emph{outcome-grounded} when its execution guidance is supported by local action-outcome evidence, irrespective of the  trajectory-level outcome $y_\tau$.
Such evidence may support either effective behavior or condition-specific avoidance guidance.
When transferred to a new task, $R_i$ must hold under the current conditions, and $P_i$ must be verified from new execution evidence.

\subsection{\nickname}
\label{sec:embodiexp}

\begin{figure*}[t]
\centering
\centerline{
\includegraphics[width=\textwidth]{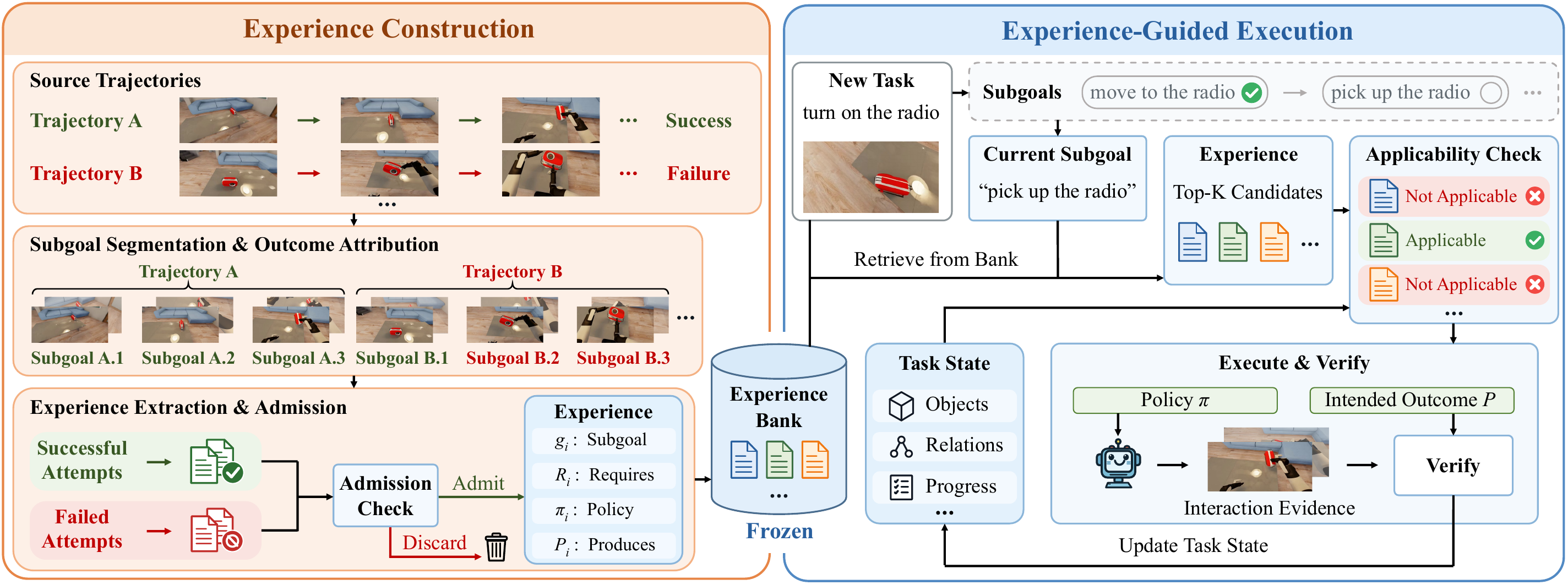}
}
\vspace{-0.3cm}
\caption{
\textbf{Overview of \nickname.}
The \emph{experience construction} stage constructs a bank for outcome-grounded subgoal experience from both successful and failed trajectories, while the \emph{experience-guided execution} stage retrieves and composes relevant experience to inform execution.
}
\label{fig:framework}
\vspace{-0.5cm}
\end{figure*}

Our framework, \textbf{\nickname}, consists of two stages: \emph{experience construction} and \emph{experience-guided execution} (Figure~\ref{fig:framework}).
During \emph{experience construction}, an experience constructor $\mathcal{C}$ processes source trajectories $\mathcal{D}_{\mathrm{src}}$.
It segments each trajectory into subgoal attempts and attributes outcomes using local action-outcome evidence.
Verified subgoal completions support guidance on effective behavior, while explicit evidence of failed operations supports condition-specific avoidance guidance.
Both are represented through the shared interface in Equation~\ref{eq:subgoal_experience}, forming an experience bank
\begin{equation}
    \mathcal{B} = \{e_i\}_{i=1}^{N},
    \label{eq:experience_bank}
\end{equation}
where $N$ is the number of admitted experience entries.
Admission is determined by whether the extracted guidance is supported by local evidence, rather than by trajectory-level success.

During \emph{experience-guided execution}, $\mathcal{B}$ remains frozen, while the executor $\mathcal{A}$ maintains an evolving \emph{Task State} $S_k$ that summarizes task-relevant facts currently supported by interaction evidence.
At each planning step $k$, $\mathcal{A}$ identifies the current subgoal and retrieves the top-$K$ relevant experiences from
$\mathcal{B}$.
An experience is used only if its \textsc{Requires} are satisfied in $S_k$ and its \textsc{Produces} is not already established.
The \textsc{Policy} of a selected experience informs grounded action generation.
After execution, newly acquired interaction evidence is used to verify the intended outcome, and update the Task State with newly supported facts before the next planning step.

This design enables two key mechanisms.
\emph{Experience Composition} supports online composition by allowing the verified outcome of one experience to satisfy the conditions required by subsequent experiences (Section~\ref{sec:composition}).
\emph{Failure Utilization} applies the same local outcome attribution to failed trajectories, retaining verified local successes and localized avoidance guidance (Section~\ref{sec:failure}).

\begin{algorithm}[t]
\caption{Online Experience Composition}
\label{alg:composition}
\begin{algorithmic}[1]
\Require Task instruction $\mathcal{I}$, frozen experience bank $\mathcal{B}$,
executor $\mathcal{A}$, environment $\mathcal{E}$

\State $o \gets \Call{Reset}{\mathcal{E}}$
\State $S_0 \gets \Call{InitializeState}{\mathcal{I},o}$
\State $k \gets 0$

\While{episode is active and budget remains}
    \State $\hat{g}_k \gets
    \Call{IdentifySubgoal}{\mathcal{I},S_k,o}$
    \State $\mathcal{K}_k \gets
    \Call{RetrieveTopK}{\mathcal{B},\mathcal{I},\hat{g}_k}$
    \State $\mathcal{U}_k \gets
    \{e \in \mathcal{K}_k \mid
    S_k \models R(e) \land S_k \not\models P(e)\}$

    \If{$\mathcal{U}_k \neq \varnothing$}
        \State $e_k \gets \Call{SelectTop}{\mathcal{U}_k}$
        \State $\hat{\mathbf{a}}_k \gets
        \Call{GenerateActions}{\mathcal{A},\mathcal{I},\hat{g}_k,S_k,o,\pi(e_k)}$
    \Else
        \State $e_k \gets \varnothing$
        \State $\hat{\mathbf{a}}_k \gets
        \Call{GenerateActions}{\mathcal{A},\mathcal{I},\hat{g}_k,S_k,o}$
    \EndIf

    \State $(o,\Omega_k^{\mathrm{pub}}) \gets
    \Call{Execute}{\mathcal{E},\hat{\mathbf{a}}_k}$

    \If{$e_k \neq \varnothing$ \textbf{and}
        $W_{\mathrm{pub}}(P(e_k),\Omega_k^{\mathrm{pub}})=1$}
        \State $S_{k+1} \gets
        \Call{UpdateState}{S_k,\Omega_k^{\mathrm{pub}},P(e_k)}$
    \Else
        \State $S_{k+1} \gets
        \Call{UpdateState}{S_k,\Omega_k^{\mathrm{pub}}}$
    \EndIf

    \State $k \gets k+1$
\EndWhile
\end{algorithmic}
\end{algorithm}

\subsubsection{Experience Composition}
\label{sec:composition}

Many embodied tasks involve multiple subgoals, whose relevant experience may originate from different source trajectories.
Rather than retrieving a fixed sequence of experiences in advance, \nickname \emph{composes} them online through the evolving Task State.
Verified outcomes update this shared state and may establish the conditions for subsequent experience.

Formally, let $e_i=(g_i,R_i,\pi_i,P_i)$ be a retrieved candidate experience for the current subgoal $\hat{g}_k$ at planning step $k$.
We write $S_k \models q$ to denote that a condition $q$ is supported by the current Task State.
The candidate can be used only if
\begin{equation}
    S_k \models R_i
    \qquad \text{and} \qquad
    S_k \not\models P_i,
    \label{eq:experience_applicability}
\end{equation}
\textit{i.e.}, its required conditions are supported, while its intended outcome is not established by the current evidence.
The executor selects the highest-ranked applicable candidate,
denoted $e_k$, and uses its policy to guide action generation.
If no candidate is applicable, $e_k=\varnothing$ and the executor
generates actions without experience guidance.

Each planning step may execute one or more native actions.
Let $\Omega_k^{\mathrm{pub}}$ denote the interaction evidence obtained over this execution span.
When an experience is selected, the executor verifies its intended outcome using Equation~\ref{eq:experience_completion} with this evidence.
It then revises the Task State by incorporating newly supported facts and updating or removing facts invalidated by the new evidence, yielding
\begin{equation}
    S_k \xrightarrow{\,e_k\,} S_{k+1},
    \qquad e_k \in \mathcal{B}\cup\{\varnothing\}.
    \label{eq:experience_transition}
\end{equation}
The intended outcome is credited only when verified. 
Other supported changes may still update the state when completion remains unverified.
At the next planning step, the executor identifies the current subgoal and retrieves candidates again, checking their applicability against $S_{k+1}$ using Equation~\ref{eq:experience_applicability}.
Repeated application of this process yields the \emph{state-grounded} composition
\begin{equation}
    S_0
    \xrightarrow{\,e_0\,}
    S_1
    \xrightarrow{\,e_1\,}
    \cdots
    \xrightarrow{\,e_{m-1}\,}
    S_m.
    \label{eq:experience_composition}
\end{equation}

Importantly, composition does not require a direct pairwise match between \textsc{Produces} and \textsc{Requires}.
The Task State integrates verified outcomes with other supported facts, allowing the conditions for a subsequent experience to be established jointly.
Thus, experience guides action generation, while verified execution outcomes support progression across subgoals (Algorithm~\ref{alg:composition}).

\subsubsection{Failure Utilization}
\label{sec:failure}

Task-level failure does not imply failure of every subgoal attempt, just as task-level success does not imply that every operation succeeded.
We therefore apply the same local outcome attribution to all source trajectories, retaining effective behavior and condition-specific
avoidance guidance.

For a source trajectory $\tau\in\mathcal{D}_{\mathrm{src}}$, let $\{\sigma_i\}_{i=1}^{M_\tau}$ denote its subgoal attempts.
Each $\sigma_i$ pursues subgoal $g_i$ with intended outcome $P_i$ and provides interaction evidence $\Omega_i^{\mathrm{pub}}$.
When $W_{\mathrm{pub}}(P_i,\Omega_i^{\mathrm{pub}})=1$, the constructor $\mathcal{C}$ abstracts successful local behavior into $\pi$ and its supported applicability conditions into $R$.
Explicit local failure evidence supports condition-specific guidance in $\pi$ on operations to avoid, with $R$ specifying the observed conditions under which those operations failed.

Both forms of guidance use the shared interface $e=(g_i,R,\pi,P_i)$ in Equation~\ref{eq:subgoal_experience}.
In both cases, $R$ specifies when the guidance applies, $\pi$ guides the pursuit of $g_i$, and $P_i$ denotes the intended completion condition rather than the observed source outcome.
The constructor $\mathcal{C}$ performs both extraction and evidence-based admission, returning the locally supported entries from each attempt:
\begin{equation}
    \mathcal{B}_i
    = \mathcal{C}\!\left(\sigma_i,\Omega_i^{\mathrm{pub}}\right).
    \label{eq:experience_extraction}
\end{equation}
A completed attempt may contribute both forms of guidance, since it may contain failed operations alongside successful behavior.
Each entry must therefore be grounded in the evidence for the relevant operations, rather than inferred from the attempt's final outcome alone.
Unverified completion alone is not evidence of failure.
If neither form of guidance is supported, $\mathcal{B}_i=\varnothing$.

The experiences are aggregated across attempts and source trajectories:
\begin{equation}
    \mathcal{B}(\tau)
    = \bigcup_{i=1}^{M_\tau}\mathcal{B}_i,
    \qquad
    \mathcal{B}
    = \bigcup_{\tau\in\mathcal{D}_{\mathrm{src}}}
      \mathcal{B}(\tau).
    \label{eq:failure_utilization}
\end{equation}
This aggregation allows failed trajectories to contribute both effective local behavior and condition-specific avoidance guidance to the experience bank.
Applicable entries guide execution, while verified outcomes update the Task State to support subsequent composition (Section~\ref{sec:composition}).

Together, these mechanisms support experience transfer across tasks: failure utilization retains effective behavior and condition-specific avoidance guidance, while composition selects and applies experience across subgoals according to the evolving Task State and verified execution outcomes.

\section{Experiments}
\label{sec:experiments}

In our experiments, we seek to answer the following questions:
\textbf{Q1:} How does \nickname compare with general-purpose LLM memory systems and embodied experience-learning methods?
\textbf{Q2:} Does subgoal-level experience transfer more effectively than trajectory-level and action-level experience?
\textbf{Q3:} How do experience composition and failure utilization contribute to experience transfer?

\textbf{Benchmarks and Metrics.} 
We evaluate on EmbodiedBench \citep{yang2025embodiedbench} and VIGIL \citep{chen2026done}.
For EmbodiedBench, we use its two environments with high-level skill interfaces: EB-Habitat for language-conditioned navigation and rearrangement in Habitat 2.0 \citep{szot2021habitat}, and EB-ALFRED for household manipulation in AI2-THOR \citep{kolve2017ai2}.
Each environment contains 300 episodes across six capability categories, with 50 episodes per category.
VIGIL complements EmbodiedBench by evaluating experience transfer over more atomic native actions.
We use its Search and Interact (SI) and Sequential Manipulation (SM) subsets, which evaluate search-conditioned interaction and ordered multi-step manipulation, respectively, with 125 episodes each.
We report task success rates, using world-state completion $W$ as the success criterion for VIGIL.

\textbf{Evaluation Protocol.}
To retain full evaluation coverage while separating experience construction from evaluation, we adopt a \emph{three-fold experience evaluation protocol} for all benchmarks.
In each rotation, one fold is \textbf{strictly held out} from experience construction, and a separate experience bank is constructed solely from trajectories in the other two folds.
The bank remains \textbf{frozen} throughout evaluation on the held-out fold.
Each episode is evaluated once, and results are aggregated over all held-out episodes across the three rotations.
All baselines follow the same fold-based protocol, with persistent memory constructed only from the construction folds and frozen during evaluation.

\textbf{Baselines.}
We compare against a no-experience baseline, general-purpose memory systems, and embodied experience-learning methods:
1) \textbf{NoExp}: The same embodied agent without persistent cross-episode experience.
2) \textbf{A-Mem}~\citep{xu2025amem}: Organizes past interactions into structured, dynamically linked memory notes.
3) \textbf{Mem0}~\citep{chhikara2025mem0}: Extracts, retrieves, and incrementally updates salient information from past interactions for long-term use.
4) \textbf{LangMem}~\citep{langmem2025}: Supports the extraction, consolidation, and retrieval of episodic and procedural memories.
5) \textbf{ReMe}~\citep{cao2025reme}: Abstracts procedural experience from historical trajectories to guide execution on new tasks.
6) \textbf{WorldMind}~\citep{ren2026worldmind}: Extracts process and goal experience into a structured world-knowledge repository.
7) \textbf{EmbodiSkill}~\citep{ju2026embodiskill}: Maintains and revises persistent procedural skills using feedback from previous trajectories.

\begin{table*}[t]
\centering
\caption{
Task Success Rates (\%) on EmbodiedBench using Qwen3.6-35B-A3B \citep{qwen36}.
Avg.\ denotes overall success rate across all 300 held-out episodes in each environment.
}
\label{tab:embodiedbench}
\setlength{\tabcolsep}{3pt} 
\renewcommand{\arraystretch}{1.1}
\resizebox{\textwidth}{!}{
\begin{tabular}{lcccccccccccccc}
\toprule
\multirow{2}{*}{\textbf{Method}}
& \multirow{2}{*}{\textbf{Avg.}}
& \multicolumn{6}{c}{\textbf{EB-Habitat}}
& \multirow{2}{*}{\textbf{Avg.}}
& \multicolumn{6}{c}{\textbf{EB-ALFRED}} \\
\cmidrule(lr){3-8}
\cmidrule(lr){10-15}
&
& \textbf{Base}
& \textbf{Com.}
& \textbf{Comp.}
& \textbf{Vis.}
& \textbf{Spa.}
& \textbf{Long}
&
& \textbf{Base}
& \textbf{Com.}
& \textbf{Comp.}
& \textbf{Vis.}
& \textbf{Spa.}
& \textbf{Long} \\
\midrule
NoExp
& 50.0 & 90 & 36 & 48 & 52 & 34 & 40
& 54.7 & 60 & 58 & 58 & 42 & 48 & 62 \\

A-Mem~\citep{xu2025amem}
& 55.0 & 94 & 48 & 54 & 56 & 36 & 42
& 59.3 & 74 & 56 & 64 & 44 & 50 & 68 \\

Mem0~\citep{chhikara2025mem0}
& 54.7 & 90 & 52 & 56 & 56 & 32 & 42
& 58.7 & 68 & 56 & 66 & 52 & 48 & 62 \\

LangMem~\citep{langmem2025}
& 58.0 & 92 & 50 & 60 & 56 & 32 & 58
& 55.0 & 58 & 52 & 66 & 46 & 38 & 70 \\

ReMe~\citep{cao2025reme}
& 53.7 & 88 & 52 & 50 & 48 & 34 & 50
& 57.3 & 64 & 54 & 66 & 46 & 44 & 70 \\

WorldMind~\citep{ren2026worldmind}
& 51.7 & 88 & 46 & 64 & 44 & 32 & 36
& 55.7 & 66 & 52 & 54 & 44 & 52 & 66 \\

EmbodiSkill~\citep{ju2026embodiskill}
& 52.3 & 90 & 36 & 64 & 44 & 36 & 44
& 51.0 & 60 & 44 & 56 & 38 & 44 & 64 \\

\textbf{EmbodiExp (Ours)}
& \textbf{68.0} & \textbf{94} & \textbf{66} & \textbf{68}
& \textbf{72} & \textbf{38} & \textbf{70}
& \textbf{68.3} & \textbf{76} & \textbf{68} & \textbf{74}
& \textbf{56} & \textbf{60} & \textbf{76} \\
\bottomrule
\end{tabular}
}
\vspace{-0.5cm}
\end{table*}

\begin{table}[t]
\centering
\caption{
World-State Completion Rates (\%) on VIGIL using Qwen3.6-27B~\citep{qwen3.6-27b}.
Avg.\ denotes the overall completion rate across all 250 held-out episodes from SI and SM.
}
\label{tab:vigil}

\scriptsize
\setlength{\tabcolsep}{3pt}
\renewcommand{\arraystretch}{1.1}

\begin{tabular*}{\textwidth}{l@{\extracolsep{\fill}\hspace{2\tabcolsep}}*{8}{c}}
\toprule
\textbf{Task}
& \textbf{NoExp}
& \textbf{A-Mem}
& \textbf{Mem0}
& \textbf{LangMem}
& \textbf{ReMe}
& \textbf{WorldMind}
& \textbf{EmbodiSkill}
& \textbf{EmbodiExp (Ours)} \\
\midrule

\textbf{SI}
& 25.6 & 32.0 & 34.4 & 32.8 & 30.4 & 28.8 & 33.6
& \textbf{36.8} \\

\textbf{SM}
& 37.6 & 38.4 & 30.4 & 36.0 & 42.4 & 40.0 & 33.6
& \textbf{44.8} \\

\midrule
\textbf{Avg.}
& 31.6 & 35.2 & 32.4 & 34.4 & 36.4 & 34.4 & 33.6
& \textbf{40.8} \\

\bottomrule
\end{tabular*}
\vspace{-0.5cm}
\end{table}

\subsection{Main Results}
\label{sec:results}

\textbf{Evaluation on EmbodiedBench.}
As shown in Table~\ref{tab:embodiedbench}, \nickname achieves overall success rates of 68.0\% on EB-Habitat and 68.3\% on EB-ALFRED, exceeding the strongest baselines by 10.0 and 9.0 percentage points, respectively.
LangMem and A-Mem provide useful execution guidance through procedural memories and linked interaction notes, achieving strongest baseline results.
However, their uneven gains across task categories and environments suggest that useful historical guidance may not remain applicable under new task conditions.
WorldMind and EmbodiSkill improve Complex tasks on EB-Habitat without comparable Long-Horizon gains, pointing to the additional difficulty of connecting useful strategies across successive subgoals.
\nickname addresses both challenges by pairing each local policy with applicability conditions and verifiable outcomes: applicability checks ground experience selection in the current state, while outcome verification grounds subsequent experience application in confirmed progress rather than assumed action success.

\begin{figure*}[t]
\centering
\centerline{
\includegraphics[width=\textwidth]{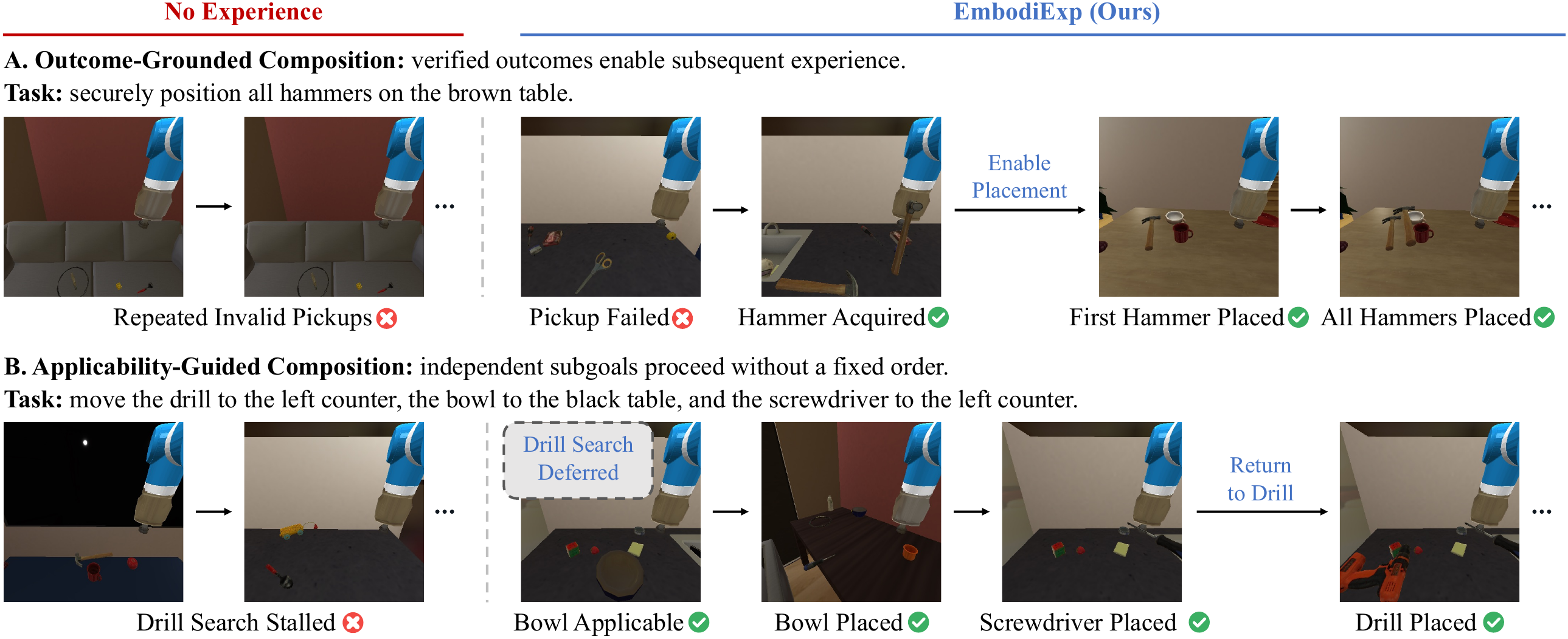}
}
\vspace{-0.3cm}
\caption{
Qualitative Comparison with NoExp on EmbodiedBench.
A. Verified hammer acquisition enables subsequent placement, while NoExp repeats invalid pickups.
B. \nickname completes applicable subgoals before returning to place the drill, while NoExp remains stalled.
}
\label{fig:qualitative}
\vspace{-0.5cm}
\end{figure*}

In Figure~\ref{fig:qualitative}.A, \nickname recovers from invalid pickup attempts, with verified hammer acquisition establishing the conditions for subsequent placement experience, whereas NoExp repeats invalid pickups.
In Figure~\ref{fig:qualitative}.B, \nickname defers the unresolved drill search and applies experience for other subgoals before returning to complete the task.
Together, these cases illustrate experience transfer through state-grounded composition: verified outcomes connect dependent subgoal experiences, while applicability allows independent ones to be reordered to suit the new task.

\textbf{Evaluation on VIGIL.}
As shown in Table~\ref{tab:vigil}, \nickname achieves an overall world-state completion rate of 40.8\%, exceeding NoExp and the strongest baseline, ReMe, by 9.2 and 4.4 percentage points, respectively.
Mem0 improves SI but regresses on SM, suggesting that memory useful for shorter interactions may not sustain longer sequences of dependent actions.
ReMe improves both subsets, suggesting that procedural experience helps multi-step execution, but guidance on how to proceed does not establish whether intermediate outcomes have actually been achieved.
\nickname instead checks applicability and connects subgoal experiences through verified outcomes, grounding progression in achieved rather than assumed effects.
These gains under VIGIL's more atomic action interface support outcome-grounded subgoals as transferable experience beyond high-level skills.













\subsection{Ablation Study}
\label{sec:ablation}

\begin{table*}[t]
\centering
\caption{
Ablation Study on EB-Habitat using Qwen3.6-35B-A3B.
Results are reported as task success rate (\%), with Avg.\ computed over all 300 held-out episodes.
}
\label{tab:ablation}

\scriptsize
\setlength{\tabcolsep}{5pt}
\renewcommand{\arraystretch}{1.1}

\begin{tabular}{lccccccc}
\toprule
\multirow{2}{*}{\textbf{Ablation}}
& \multirow{2}{*}{\textbf{Avg.}}
& \multicolumn{6}{c}{\textbf{EB-Habitat}} \\
\cmidrule(lr){3-8}
&
& \textbf{Base}
& \textbf{Com.}
& \textbf{Comp.}
& \textbf{Vis.}
& \textbf{Spa.}
& \textbf{Long} \\
\midrule

NoExp
& 50.0 & 90 & 36 & 48 & 52 & 34 & 40 \\

\midrule

\multicolumn{8}{l}{\textit{Experience Granularity}} \\

Trajectory-Level Experience
& 54.0 & 88 & 58 & 46 & 58 & \textbf{38} & 36 \\

Action-Level Experience
& 54.7 & 92 & 48 & 54 & 58 & 36 & 40 \\

Subgoal-Level Experience
& 58.3 & 90 & 60 & 56 & 60 & \textbf{38} & 46 \\

\textbf{Full \nickname (Ours)}
& \textbf{68.0} & \textbf{94} & \textbf{66} & \textbf{68}
& \textbf{72} & \textbf{38} & \textbf{70} \\

\midrule

\multicolumn{8}{l}{\textit{Experience Composition}} \\

Similarity-Based Composition
& 58.3 & 90 & 60 & 56 & 60 & \textbf{38} & 46 \\

Applicability Checks Only
& 64.3 & 92 & \textbf{66} & 64 & 68 & \textbf{38} & 58 \\

Outcome Verification Only
& 56.7 & 88 & 60 & 56 & 60 & 36 & 40 \\

\textbf{State-Grounded Composition (Ours)}
& \textbf{68.0} & \textbf{94} & \textbf{66} & \textbf{68}
& \textbf{72} & \textbf{38} & \textbf{70} \\

\midrule

\multicolumn{8}{l}{\textit{Failure Utilization}} \\

Successful Trajectories Only
& 62.7 & 92 & \textbf{66} & 62 & 64 & \textbf{38} & 54 \\

\textbf{Successful + Failed Trajectories (Ours)}
& \textbf{68.0} & \textbf{94} & \textbf{66} & \textbf{68}
& \textbf{72} & \textbf{38} & \textbf{70} \\









\bottomrule
\end{tabular}
\vspace{-0.5cm}
\end{table*}


\textbf{Experience Granularity.}
We compare trajectory-, action-, and subgoal-level experience under the same similarity-based top-1 retrieval protocol, without applicability checks or outcome verification (Table~\ref{tab:ablation}).
Subgoal-level experience achieves 58.3\% success, outperforming trajectory-level (54.0\%) and action-level (54.7\%) alternatives.
The difference is more pronounced on Long-Horizon tasks, where trajectory- and action-level experience provide no gain over NoExp, while subgoal-level experience improves success from 40\% to 46\%.
This suggests that subgoal-level experience strikes a better balance for cross-task transfer by preserving the action context needed for a local outcome while avoiding source-task-specific dependencies carried by full trajectories.

\textbf{Experience Composition.}
Similarity-Based Composition uses semantic top-1 retrieval without applicability checks or outcome verification, achieving 58.3\% success (Table~\ref{tab:ablation}).
Applicability Only improves success to 64.3\%, whereas Outcome Verification Only reaches 56.7\%.
Combining both components raises success to 68.0\%, with the largest gain on Long-Horizon tasks.
This contrast indicates that outcome verification is most effective when coupled with applicability-aware selection: applicability identifies experience suited to the current state, while verification determines whether its execution actually establishes the conditions needed for subsequent experience.
Together, they ground successive subgoal transitions in achieved outcomes rather than assumed progress.

\textbf{Failure Utilization.}
Including failed source trajectories raises success from 62.7\% to 68.0\% (Table~\ref{tab:ablation}).
This shows that failed trajectories can still contain valuable local experience: verified local successes can provide transferable policies, while localized failures identify context-specific actions that should be avoided.
Through subgoal-level outcome attribution, \nickname retains both forms of evidence instead of discarding them based on the final task outcome.
The largest gain on Long-Horizon tasks (54\% to 70\%) suggests that such localized evidence is especially useful in extended tasks, both to avoid unsuccessful actions and to compose successful subgoal transitions.








\begin{figure*}[t]
\centering
\centerline{
\includegraphics[width=\textwidth]{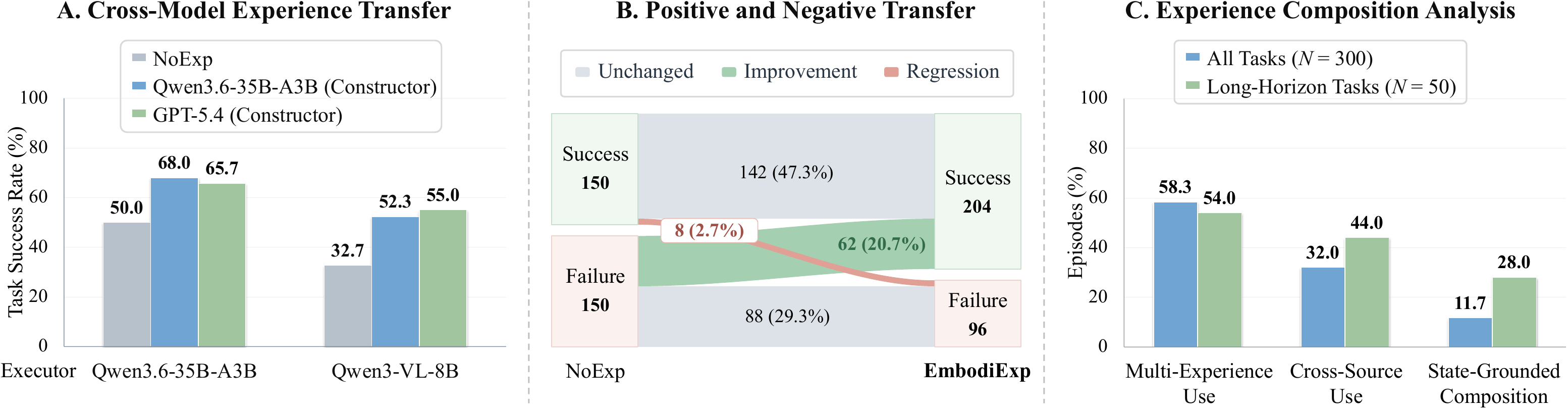}
}
\vspace{-0.3cm}
\caption{
Additional Analysis on EB-Habitat.
A. Task success rates with different experience constructors and executors.
B. Paired outcome transitions from NoExp to \nickname, with counts and percentages over all 300 episodes.
C. Three nested experience-use criteria for all tasks ($N=300$) and Long-Horizon subset ($N=50$), with percentages computed over all episodes in each set.
}
\label{fig:analysis}
\vspace{-0.5cm}
\end{figure*}

\subsection{Additional Analysis}
\label{sec:analysis}

\textbf{Cross-Model Experience Transfer.}
Both banks improve both executors, enabling the 8B \citep{bai2025qwen3vl} executor reaching 52.3-55.0\% success, above the 35B NoExp baseline (Figure~\ref{fig:analysis}.A).
These gains suggest that transferred experience can partially compensate for limited executor capability without increasing executors' model size.
The constructor ranking nevertheless reverses across executors: the Qwen bank performs better with 35B, whereas the GPT \citep{openai2026gpt54} bank performs better with 8B.
Together, these results support separating offline experience construction from online execution, while indicating that the utility of a bank depends on the constructor-executor pairing.

\textbf{Positive and Negative Transfer.}
Across 300 paired episodes (Figure~\ref{fig:analysis}.B), \nickname converts 62 NoExp failures into successes (20.7\%), while 8 successes regress (2.7\%), yielding a net gain of 18.0 percentage points.
This corresponds to recovering 41.3\% of baseline failures while preserving 94.7\% of baseline successes.
The gains therefore come mainly from solving tasks that NoExp fails, while preserving most baseline successes.
However, the remaining regressions show that experience transfer can still reduce performance on individual tasks despite improving overall success.

\textbf{Experience Composition Analysis.}
Figure~\ref{fig:analysis}.C reports three nested criteria: multiple experience applications, use across disjoint source-trajectory sets, and state-grounded cross-source composition.
The last criterion requires verified local facts from earlier applications to remain valid in the Task State and support later prerequisites.
Long-Horizon tasks show slightly less multi-experience use than overall, but more cross-source use and state-grounded composition, with the latter occurring in 28.0\% of Long-Horizon episodes versus 11.7\% overall.
These traces provide execution-level evidence of compositional experience transfer: verified outcomes from experience drawn from one source establish prerequisites for applying experience drawn from another.

\section{Conclusion}

In this paper, we present \textbf{\nickname}, a framework for constructing transferable experiences and composing them in embodied tasks.
At its core is \emph{outcome-grounded subgoal experience}, which specifies a subgoal, applicability conditions, an adaptable policy, and an intended local outcome.
\nickname composes experiences across trajectories through verified outcomes that make subsequent experiences applicable, while subgoal-level outcome attribution derives execution and avoidance guidance from local successes and failures.
Fold-based experiments on multiple embodied benchmarks with frozen experience banks demonstrate its effectiveness.
These findings underscore the importance of applicability checks and outcome verification for compositional transfer.

\subsubsection*{Acknowledgments}
This work is supported in part by the Natural Science Foundation of China
(42671592, 42301520), the Research Grants Council of Hong Kong
(25206524, 15212925).

\bibliography{main}

@article{ahn2022saycan,
  title   = {{Do As I Can, Not As I Say: Grounding Language in Robotic Affordances}},
  author  = {Ahn, Michael and Brohan, Anthony and Brown, Noah and Chebotar, Yevgen and Cortes, Omar and David, Byron and Finn, Chelsea and Fu, Chuyuan and Gopalakrishnan, Keerthana and Hausman, Karol and others},
  journal = {arXiv preprint arXiv:2204.01691},
  year    = {2022}
}

@inproceedings{liang2023codeaspolicies,
  title     = {{Code as Policies: Language Model Programs for Embodied Control}},
  author    = {Liang, Jacky and Huang, Wenlong and Xia, Fei and Xu, Peng and Hausman, Karol and Ichter, Brian and Florence, Pete and Zeng, Andy},
  booktitle = {ICRA},
  year      = {2023}
}

@inproceedings{singh2023progprompt,
  title     = {{ProgPrompt: Generating Situated Robot Task Plans using Large Language Models}},
  author    = {Singh, Ishika and Blukis, Valts and Mousavian, Arsalan and Goyal, Ankit and Xu, Danfei and Tremblay, Jonathan and Fox, Dieter and Thomason, Jesse and Garg, Animesh},
  booktitle = {ICRA},
  year      = {2023}
}

@inproceedings{huang2023innermonologue,
  title     = {{Inner Monologue: Embodied Reasoning through Planning with Language Models}},
  author    = {Huang, Wenlong and Xia, Fei and Xiao, Ted and Chan, Harris and Liang, Jacky and Florence, Pete and Zeng, Andy and Tompson, Jonathan and Mordatch, Igor and Chebotar, Yevgen and others},
  booktitle = {CoRL},
  year      = {2022}
}

@inproceedings{song2023llmplanner,
  title     = {{LLM-Planner: Few-Shot Grounded Planning for Embodied Agents with Large Language Models}},
  author    = {Song, Chan Hee and Wu, Jiaman and Washington, Clayton and Sadler, Brian M. and Chao, Wei-Lun and Su, Yu},
  booktitle = {ICCV},
  year      = {2023}
}

@inproceedings{wang2023deps,
  title     = {{Describe, Explain, Plan and Select: Interactive Planning with Large Language Models Enables Open-World Multi-Task Agents}},
  author    = {Wang, Zihao and Cai, Shaofei and Chen, Guanzhou and Liu, Anji and Ma, Xiaojian and Liang, Yitao},
  booktitle = {NeurIPS},
  year      = {2023}
}

@article{wang2023voyager,
  title   = {{Voyager: An Open-Ended Embodied Agent with Large Language Models}},
  author  = {Wang, Guanzhi and Xie, Yuqi and Jiang, Yunfan and Mandlekar, Ajay and Xiao, Chaowei and Zhu, Yuke and Fan, Linxi and Anandkumar, Anima},
  journal = {arXiv preprint arXiv:2305.16291},
  year    = {2023}
}

@article{driess2023palme,
  title     = {{PaLM-E: An Embodied Multimodal Language Model}},
  author    = {Driess, Danny and Xia, Fei and Sajjadi, Mehdi S. M. and Lynch, Corey and Chowdhery, Aakanksha and Ichter, Brian and Wahid, Ayzaan and Tompson, Jonathan and Vuong, Quan and Yu, Tianhe and others},
  journal   = {arXiv preprint arXiv:2303.03378},
  year      = {2023}
}

@article{brohan2023rt2,
  title     = {{RT-2: Vision-Language-Action Models Transfer Web Knowledge to Robotic Control}},
  author    = {Brohan, Anthony and Brown, Noah and Carbajal, Justice and Chebotar, Yevgen and Chen, Xi and Choromanski, Krzysztof and Ding, Tianli and Driess, Danny and Dubey, Avinava and Finn, Chelsea and others},
  journal   = {arXiv preprint arXiv:2307.15818},
  year      = {2023}
}

@article{zitkovich2023rtx,
  title     = {{Open X-Embodiment: Robotic Learning Datasets and RT-X Models}},
  author    = {{Open X-Embodiment Collaboration}},
  journal   = {arXiv preprint arXiv:2310.08864},
  year      = {2023}
}

@inproceedings{kim2024openvla,
  title     = {{OpenVLA: An Open-Source Vision-Language-Action Model}},
  author    = {Kim, Moo Jin and Pertsch, Karl and Karamcheti, Siddharth and Xiao, Ted and Balakrishna, Ashwin and Nair, Suraj and Rafailov, Rafael and Foster, Ethan P. and Sanketi, Pannag R. and Vuong, Quan and others},
  booktitle = {CoRL},
  year      = {2024}
}

@inproceedings{octo2024,
  title     = {{Octo: An Open-Source Generalist Robot Policy}},
  author    = {{Octo Model Team} and Ghosh, Dibya and Walke, Homer and Pertsch, Karl and Black, Kevin and Mees, Oier and Dasari, Sudeep and Hejna, Joey and Kreiman, Tobias and Xu, Charles and others},
  booktitle = {Robotics: Science and Systems},
  year      = {2024}
}

@inproceedings{shridhar2020alfred,
  title     = {{ALFRED: A Benchmark for Interpreting Grounded Instructions for Everyday Tasks}},
  author    = {Shridhar, Mohit and Thomason, Jesse and Gordon, Daniel and Bisk, Yonatan and Han, Winson and Mottaghi, Roozbeh and Zettlemoyer, Luke and Fox, Dieter},
  booktitle = {CVPR},
  year      = {2020}
}

@inproceedings{shridhar2021alfworld,
  title     = {{ALFWorld: Aligning Text and Embodied Environments for Interactive Learning}},
  author    = {Shridhar, Mohit and Yuan, Xingdi and Cote, Marc-Alexandre and Bisk, Yonatan and Trischler, Adam and Hausknecht, Matthew},
  booktitle = {ICLR},
  year      = {2021}
}

@inproceedings{yang2025embodiedbench,
  title     = {{EmbodiedBench: Comprehensive Benchmarking Multi-modal Large Language Models for Vision-Driven Embodied Agents}},
  author    = {Yang, Rui and Chen, Hanyang and Zhang, Junyu and Zhao, Mark and Qian, Cheng and Wang, Kangrui and Wang, Qineng and Koripella, Teja Venkat and Movahedi, Marziyeh and Li, Manling and others},
  booktitle = {ICML},
  year      = {2025}
}

@article{park2023generative,
  title   = {{Generative Agents: Interactive Simulacra of Human Behavior}},
  author  = {Park, Joon Sung and O'Brien, Joseph C. and Cai, Carrie J. and Morris, Meredith Ringel and Liang, Percy and Bernstein, Michael S.},
  journal = {arXiv preprint arXiv:2304.03442},
  year    = {2023}
}

@article{packer2023memgpt,
  title   = {{MemGPT: Towards LLMs as Operating Systems}},
  author  = {Packer, Charles and Wooders, Sarah and Lin, Kevin and Fang, Vivian and Patil, Shishir G. and Stoica, Ion and Gonzalez, Joseph E.},
  journal = {arXiv preprint arXiv:2310.08560},
  year    = {2023}
}

@inproceedings{zhong2024memorybank,
  title     = {{MemoryBank: Enhancing Large Language Models with Long-Term Memory}},
  author    = {Zhong, Wanjun and Guo, Lianghong and Gao, Qiqi and Ye, He and Wang, Yanlin},
  booktitle = {AAAI},
  year      = {2024}
}

@inproceedings{shinn2023reflexion,
  title     = {{Reflexion: Language Agents with Verbal Reinforcement Learning}},
  author    = {Shinn, Noah and Cassano, Federico and Gopinath, Ashwin and Narasimhan, Karthik R. and Yao, Shunyu},
  booktitle = {NeurIPS},
  year      = {2023}
}

@article{zhao2024expel,
  title   = {{ExpeL: LLM Agents Are Experiential Learners}},
  author  = {Zhao, Andrew and Huang, Daniel and Xu, Quentin and Lin, Matthieu and Liu, Yong-Jin and Huang, Gao},
  journal = {arXiv preprint arXiv:2308.10144},
  year    = {2024}
}

@article{chhikara2025mem0,
  title   = {{Mem0: Building Production-Ready AI Agents with Scalable Long-Term Memory}},
  author  = {Chhikara, Prateek and Khant, Dev and Aryan, Saket and Singh, Taranjeet and Yadav, Deshraj},
  journal = {arXiv preprint arXiv:2504.19413},
  year    = {2025}
}

@article{xu2025amem,
  title   = {{A-MEM: Agentic Memory for LLM Agents}},
  author  = {Xu, Wujiang and Liang, Zujie and Mei, Kai and Gao, Hang and Tan, Juntao and Zhang, Yongfeng},
  journal = {arXiv preprint arXiv:2502.12110},
  year    = {2025}
}

@article{zhang2025gmemory,
  title   = {G{-Memory: Tracing Hierarchical Memory for Multi-Agent Systems}},
  author  = {Zhang, Guibin and Fu, Muxin and Wan, Guancheng and Yu, Miao and Wang, Kun and Yan, Shuicheng},
  journal = {arXiv preprint arXiv:2506.07398},
  year    = {2025}
}

@article{wang2024awm,
  title   = {{Agent Workflow Memory}},
  author  = {Wang, Zora Zhiruo and Mao, Jiayuan and Fried, Daniel and Neubig, Graham},
  journal = {arXiv preprint arXiv:2409.07429},
  year    = {2024}
}

@inproceedings{fang2026memp,
  title     = {{Memp: Exploring Agent Procedural Memory}},
  author    = {Fang, Runnan and Liang, Yuan and Wang, Xiaobin and Wu, Jialong and Qiao, Shuofei and Xie, Pengjun and Huang, Fei and Chen, Huajun and Zhang, Ningyu},
  booktitle = {ACL Findings},
  year      = {2026}
}

@article{wang2024karma,
  title   = {{KARMA: Augmenting Embodied AI Agents with Long-and-Short Term Memory Systems}},
  author  = {Wang, Zixuan and Yu, Bo and Zhao, Junzhe and Sun, Wenhao and Hou, Sai and Liang, Shuai and Hu, Xing and Han, Yinhe and Gan, Yiming},
  journal = {arXiv preprint arXiv:2409.14908},
  year    = {2024}
}

@inproceedings{yang2025threedmem,
  title     = {{3D-Mem: 3D Scene Memory for Embodied Exploration and Reasoning}},
  author    = {Yang, Yuncong and Yang, Han and Zhou, Jiachen and Chen, Peihao and Zhang, Hongxin and Du, Yilun and Gan, Chuang},
  booktitle = {CVPR},
  year      = {2025}
}

@article{hu2025threedllmmem,
  title   = {{3DLLM-Mem: Long-Term Spatial-Temporal Memory for Embodied 3D Large Language Model}},
  author  = {Hu, Wenbo and Hong, Yining and Wang, Yanjun and Gao, Leison and Wei, Zibu and Yao, Xingcheng and Peng, Nanyun and Bitton, Yonatan and Szpektor, Idan and Chang, Kai-Wei},
  journal = {arXiv preprint arXiv:2505.22657},
  year    = {2025}
}

@article{ma2026brainmem,
  title   = {{BrainMem: Brain-Inspired Evolving Memory for Embodied Agent Task Planning}},
  author  = {Ma, Xiaoyu and Hu, Lianyu and Tang, Wenbing and Hu, Zixuan and Liao, Zeqin and Wu, Zhizhen and Liu, Yang},
  journal = {arXiv preprint arXiv:2604.16331},
  year    = {2026}
}

@article{torne2026mem,
  title   = {{MEM: Multi-Scale Embodied Memory for Vision Language Action Models}},
  author  = {Torne, Marcel and Pertsch, Karl and Walke, Homer and Vedder, Kyle and Nair, Suraj and Ichter, Brian and Ren, Allen Z. and Wang, Haohuan and Tang, Jiaming and Stachowicz, Kyle and others},
  journal = {arXiv preprint arXiv:2603.03596},
  year    = {2026}
}

@article{wang2026lmee,
  title   = {{Explore with Long-term Memory: A Benchmark and Multimodal LLM-based Reinforcement Learning Framework for Embodied Exploration}},
  author  = {Wang, Sen and Liu, Bangwei and Gao, Zhenkun and Ma, Lizhuang and Wang, Xuhong and Xie, Yuan and Tan, Xin},
  journal = {arXiv preprint arXiv:2601.10744},
  year    = {2026}
}

@article{liu2025memverse,
  title   = {{MemVerse: Multimodal Memory for Lifelong Learning Agents}},
  author  = {Liu, Junming and Sun, Yifei and Cheng, Weihua and Lei, Haodong and Chen, Yirong and Wen, Licheng and Yang, Xuemeng and Fu, Daocheng and Cai, Pinlong and Deng, Nianchen and others},
  journal = {arXiv preprint arXiv:2512.03627},
  year    = {2025}
}

@article{tang2025agentkb,
  title   = {{Agent KB: Leveraging Cross-Domain Experience for Agentic Problem Solving}},
  author  = {Tang, Xiangru and Qin, Tianrui and Peng, Tianhao and Zhou, Ziyang and Shao, Daniel and Du, Tingting and Wei, Xinming and Xia, Peng and Wu, Fang and Zhu, He and others},
  journal = {arXiv preprint arXiv:2507.06229},
  year    = {2025}
}

@article{ouyang2025reasoningbank,
  title   = {{ReasoningBank: Scaling Agent Self-Evolving with Reasoning Memory}},
  author  = {Ouyang, Siru and Yan, Jun and Hsu, I-Hung and Chen, Yanfei and Jiang, Ke and Wang, Zifeng and Han, Rujun and Le, Long T. and Daruki, Samira and Tang, Xiangru and others},
  journal = {arXiv preprint arXiv:2509.25140},
  year    = {2025}
}

@article{cao2025reme,
  title   = {{Remember Me, Refine Me: A Dynamic Procedural Memory Framework for Experience-Driven Agent Evolution}},
  author  = {Cao, Zouying and Deng, Jiaji and Yu, Li and Zhou, Weikang and Liu, Zhaoyang and Ding, Bolin and Zhao, Hai},
  journal = {arXiv preprint arXiv:2512.10696},
  year    = {2025}
}

@article{zhang2026liveevo,
  title   = {{Live-Evo: Online Evolution of Agentic Memory from Continuous Feedback}},
  author  = {Zhang, Yaolun and Wu, Yiran and Yu, Yijiong and Wu, Qingyun and Wang, Huazheng},
  journal = {arXiv preprint arXiv:2602.02369},
  year    = {2026}
}

@article{zhang2026memskill,
  title   = {{MemSkill: Learning and Evolving Memory Skills for Self-Evolving Agents}},
  author  = {Zhang, Haozhen and Long, Quanyu and Bao, Jianzhu and Feng, Tao and Zhang, Weizhi and Yue, Haodong and Wang, Wenya},
  journal = {arXiv preprint arXiv:2602.02474},
  year    = {2026}
}

@article{xia2026skillrl,
  title   = {{SkillRL: Evolving Agents via Recursive Skill-Augmented Reinforcement Learning}},
  author  = {Xia, Peng and Chen, Jianwen and Wang, Hanyang and Liu, Jiaqi and Zeng, Kaide and Wang, Yu and Han, Siwei and Zhou, Yiyang and Zhao, Xujiang and Chen, Haifeng and others},
  journal = {arXiv preprint arXiv:2602.08234},
  year    = {2026}
}

@article{cai2025flex,
  title   = {{FLEX: Continuous Agent Evolution via Forward Learning from Experience}},
  author  = {Cai, Zhicheng and Guo, Xinyuan and Pei, Yu and Feng, Jiangtao and Su, Jinsong and Chen, Jiangjie and Zhang, Ya-Qin and Ma, Wei-Ying and Wang, Mingxuan and Zhou, Hao},
  journal = {arXiv preprint arXiv:2511.06449},
  year    = {2025}
}

@article{ni2026trace2skill,
  title   = {{Trace2Skill: Distill Trajectory-Local Lessons into Transferable Agent Skills}},
  author  = {Ni, Jingwei and Liu, Yihao and Liu, Xinpeng and Sun, Yutao and Zhou, Mengyu and Cheng, Pengyu and Wang, Dexin and Zhao, Erchao and Jiang, Xiaoxi and Jiang, Guanjun},
  journal = {arXiv preprint arXiv:2603.25158},
  year    = {2026}
}

@article{zhang2026coevoskills,
  title   = {{CoEvoSkills: Self-Evolving Agent Skills via Co-Evolutionary Verification}},
  author  = {Zhang, Hanrong and Fan, Shicheng and Zou, Henry Peng and Chen, Yankai and Wang, Zhenting and Zhou, Jiayu and Li, Chengze and Huang, Wei-Chieh and Yao, Yifei and Zheng, Kening and others},
  journal = {arXiv preprint arXiv:2604.01687},
  year    = {2026}
}

@article{yang2026autoskill,
  title   = {{AutoSkill: Experience-Driven Lifelong Learning via Skill Self-Evolution}},
  author  = {Yang, Yutao and Li, Junsong and Pan, Qianjun and Zhan, Bihao and Cai, Yuxuan and Du, Lin and Zhou, Jie and Chen, Kai and Chen, Qin and Li, Xin and others},
  journal = {arXiv preprint arXiv:2603.01145},
  year    = {2026}
}

@article{zhang2026skill1,
  title   = {{Skill1: Unified Evolution of Skill-Augmented Agents via Reinforcement Learning}},
  author  = {Shi, Yaorui and Chen, Yuxin and Lu, Zhengxi and Miao, Yuchun and Liu, Shugui and Gu, Qi and Cai, Xunliang and Wang, Xiang and Zhang, An},
  journal = {arXiv preprint arXiv:2605.06130},
  year    = {2026}
}

@article{wang2026museagent,
  title   = {{MuSEAgent: A Multimodal Reasoning Agent with Stateful Experiences}},
  author  = {Wang, Shijian and Jin, Jiarui and Fu, Runhao and Yan, Zexuan and Wang, Xingjian and Hu, Mengkang and Wang, Eric and Li, Xiaoxi and Zhang, Kangning and Yao, Li and others},
  journal = {arXiv preprint arXiv:2603.27813},
  year    = {2026}
}

@inproceedings{jiang2026xskill,
  title     = {{XSkill: Continual Learning from Experience and Skills in Multimodal Agents}},
  author    = {Jiang, Guanyu and Su, Zhaochen and Qu, Xiaoye and Fung, Yi R.},
  booktitle = {ICML},
  year      = {2026}
}

@inproceedings{ren2026worldmind,
  title   = {{Aligning Agentic World Models via Knowledgeable Experience Learning}},
  author  = {Ren, Baochang and Yao, Yunzhi and Sun, Rui and Qiao, Shuofei and Zhang, Ningyu and Chen, Huajun},
  booktitle = {EMNLP},
  year    = {2026}
}

@article{ju2026embodiskill,
  title   = {{EmbodiSkill: Skill-Aware Reflection for Self-Evolving Embodied Agents}},
  author  = {Ju, Ruofei and Wang, Xinrui and Ding, Xin and Yang, Yifan and Wu, Hao and Jiang, Shiqi and Zhang, Qianxi and Wen, Hao and Li, Xiangyu and Wang, Weijun and others},
  journal = {arXiv preprint arXiv:2605.10332},
  year    = {2026}
}

@inproceedings{chen2026done,
  title   = {{Done, But Not Sure: Disentangling World Completion from Self-Termination in Embodied Agents}},
  author  = {Chen, Ying and Fang, Lihuang and Jiang, Rui and Wang, Mingxu and Gu, Zhifeng and Yi, Lei and Chen, Jie},
  booktitle = {NeurIPS},
  year    = {2026}
}

@misc{langmem2025,
  author       = {{LangChain Team}},
  title        = {{LangMem SDK for Agent Long-Term Memory}},
  year         = {2025},
  month        = feb,
  howpublished = {\url{https://www.langchain.com/blog/langmem-sdk-launch}},
}

@inproceedings{szot2021habitat,
  title={{Habitat 2.0: Training Home Assistants to Rearrange Their Habitat}},
  author={Szot, Andrew and Clegg, Alexander and Undersander, Eric and Wijmans, Erik and Zhao, Yili and Turner, John and Maestre, Noah and Mukadam, Mustafa and Chaplot, Devendra Singh and Maksymets, Oleksandr and others},
  booktitle={NeurIPS},
  year={2021}
}

@article{kolve2017ai2,
  title={{AI2-THOR: An Interactive 3D Environment for Visual AI}},
  author={Kolve, Eric and Mottaghi, Roozbeh and Han, Winson and VanderBilt, Eli and Weihs, Luca and Herrasti, Alvaro and Deitke, Matt and Ehsani, Kiana and Gordon, Daniel and Zhu, Yuke and others},
  journal={arXiv preprint arXiv:1712.05474},
  year={2017}
}

@misc{qwen36,
  title  = {{Qwen3.6-35B-A3B: Agentic Coding Power, Now Open to All}},
  author = {{Qwen Team}},
  year   = {2026},
  url    = {https://qwen.ai/blog?id=qwen3.6-35b-a3b}
}

@misc{qwen3.6-27b,
  title  = {{Qwen3.6-27B: Flagship-Level Coding in a 27B Dense Model}},
  author = {{Qwen Team}},
  year   = {2026},
  url    = {https://qwen.ai/blog?id=qwen3.6-27b}
}

@article{team2025gemini,
  title={{Gemini Robotics: Bringing AI into the Physical World}},
  author={Team, Gemini Robotics and Abeyruwan, Saminda and Ainslie, Joshua and Alayrac, Jean-Baptiste and Arenas, Montserrat Gonzalez and Armstrong, Travis and Balakrishna, Ashwin and Baruch, Robert and Bauza, Maria and Blokzijl, Michiel and others},
  journal={arXiv preprint arXiv:2503.20020},
  year={2025}
}

@inproceedings{agarwal2025l3m+,
  title={{L3M+P: Lifelong Planning with Large Language Models}},
  author={Agarwal, Krish and Jiang, Yuqian and Hu, Jiaheng and Liu, Bo and Stone, Peter},
  booktitle={IROS},
  year={2025}
}

@misc{openai2026gpt54,
  title        = {{GPT-5.4 Thinking System Card}},
  author       = {{OpenAI}},
  year         = {2026},
  howpublished = {OpenAI},
  note         = {Published March 5, 2026}
}

@inproceedings{wang2026trajrag,
  title={{TrajRAG: Retrieving Geometric-Semantic Experience for Zero-Shot Object Navigation}},
  author={Wang, Yiyao and Zhang, Sixian and Zhang, Keming and Song, Xinhang and Du, Songjie and Jiang, Shuqiang},
  booktitle={CVPR},
  year={2026}
}

@inproceedings{yue2025mllm,
  title={{MLLM as Retriever: Interactively Learning Multimodal Retrieval for Embodied Agents}},
  author={Yue, Junpeng and Xu, Xinrun and Karlsson, B{\"o}rje F and Lu, Zongqing},
  booktitle={ICLR},
  year={2025}
}

@article{konidaris2018skills,
  title={{From skills to symbols: Learning symbolic representations for abstract high-level planning}},
  author={Konidaris, George and Kaelbling, Leslie Pack and Lozano-Perez, Tomas},
  journal={Journal of Artificial Intelligence Research},
  volume={61},
  pages={215--289},
  year={2018}
}

@inproceedings{james2020learning,
  title={{Learning Portable Representations for High-Level Planning}},
  author={James, Steven and Rosman, Benjamin and Konidaris, George},
  booktitle={ICML},
  year={2020}
}

@inproceedings{konidaris2007building,
  title={{Building Portable Options: Skill Transfer in Reinforcement Learning}},
  author={Konidaris, George Dimitri and Barto, Andrew G},
  booktitle={IJCAI},
  year={2007}
}

@inproceedings{li2024embodied,
  title={{Embodied Agent Interface: Benchmarking LLMs for Embodied Decision Making}},
  author={Li, Manling and Zhao, Shiyu and Wang, Qineng and Wang, Kangrui and Zhou, Yu and Srivastava, Sanjana and Gokmen, Cem and Lee, Tony and Li, Li E and Zhang, Ruohan and others},
  booktitle={NeurIPS},
  year={2024}
}

@inproceedings{hu2025hiagent,
  title={{HiAgent: Hierarchical Working Memory Management for Solving Long-Horizon Agent Tasks with Large Language Model}},
  author={Hu, Mengkang and Chen, Tianxing and Chen, Qiguang and Mu, Yao and Shao, Wenqi and Luo, Ping},
  booktitle={ACL},
  year={2025}
}

@article{banwasi2026messymem,
  title = {{MessyMem: Learning-from-Doing Memory for Mobile Manipulation}},
  author = {Banwasi, Anuva and Muckelroy, III, William and Sundaresan, Priya and Zhao, Linfeng and Bohg, Jeannette and Ho, Cherie},
  journal={arXiv preprint arXiv:2609.15976},
  year = {2026}
}

@article{tang2026msce,
  title = {{From Memory to Skills: Evidence-Grounded Co-Evolution Governance for Long-Horizon LLM Agents}},
  author = {Tang, Bo and Zhang, Yang and Zhuang, Guomian and Wei, Wenqiang and Zheng, Gaoyang and Xie, Lindong and Tan, Yanchao and Xiong, Feiyu and Yang, Qingyu and Chung, Edward and Li, Zhiyu},
  journal={arXiv preprint arXiv:2607.16621},
  year = {2026}
}

@article{sun2026trace2tower,
  title = {{Trace2Tower: Transition-Aware EigenTrace Induction of Multi-Level Skills for LLM Agents}},
  author = {Sun, Jiazheng and Yang, Boyu and Yuan, Binhao and Li, Mingxuan and Peng, Xin},
  journal={arXiv preprint arXiv:2609.05261},
  year = {2026}
}

@article{feng2026break,
  title = {{Break It Down, Pass It On: Cross-Task Skill Transfer in LLM Agents}},
  author = {Feng, Yiyang and Bijoy, Biddut Sarker and Balasubramanian, Niranjan and Zhou, Jiawei},
  journal={arXiv preprint arXiv:2608.20274},
  year = {2026}
}

@article{wei2026elite,
  title = {{ELITE: Experiential Learning and Intent-Aware Transfer for Self-improving Embodied Agents}},
  author = {Wei, Bingqing and Xia, Zhongyu and Liu, Dingai and Zhou, Xiaoyu and Lin, Zhiwei and Wang, Yongtao},
  journal={arXiv preprint arXiv:2603.24018},
  year = {2026}
}

@article{xie2026mineevolve,
  title = {{MineEvolve: Self-Evolution with Accumulated Knowledge for Long-Horizon Embodied Minecraft Agents}},
  author = {Xie, Zhengwei and Chen, Zhisheng and Weng, Ziyan and Li, Jinhan and Li, Chenglong and Xiao, Zikai and Song, Jingwei and Jing, Jinhao and Zhang, Vireo and Wang, Kun},
  journal={arXiv preprint arXiv:2603.13131},
  year = {2026}
}

@article{chen2026capek,
  title={{Capek 0.5: An Execution-Centric Vision-Language Model for Embodied Intelligence}},
  author={Chen, Ying and Li, Weizhen and Hu, Zhe and Li, Zhenjiang and Jiang, Rui and Gu, Zhifeng and Fang, Lihuang and Liu, Jiangping and Yi, Lei and Chen, Jie},
  journal={arXiv preprint arXiv:2608.06756},
  year={2026}
}

@article{hong2026esi,
  title={{ESI-Bench: Towards Embodied Spatial Intelligence that Closes the Perception-Action Loop}},
  author={Hong, Yining and Liu, Jiageng and Yin, Han and Li, Manling and Guibas, Leonidas and Fei-Fei, Li and Wu, Jiajun and Choi, Yejin},
  journal={arXiv preprint arXiv:2605.18746},
  year={2026}
}

@article{bai2025qwen3vl,
  title   = {{Qwen3-VL Technical Report}},
  author  = {Bai, Shuai and Cai, Yuxuan and Chen, Ruizhe and others},
  journal = {arXiv preprint arXiv:2511.21631},
  year    = {2025}
}

@inproceedings{reimers2019sentencebert,
  title={{Sentence-BERT: Sentence Embeddings using Siamese BERT-Networks}},
  author={Reimers, Nils and Gurevych, Iryna},
  booktitle={EMNLP},
  year={2019}
}

@inproceedings{wang2020minilm,
  title={{MiniLM: Deep Self-Attention Distillation for Task-Agnostic Compression of Pre-Trained Transformers}},
  author={Wang, Wenhui and Wei, Furu and Dong, Li and Bao, Hangbo and Yang, Nan and Zhou, Ming},
  booktitle={NeurIPS},
  year={2020}
}
\bibliographystyle{iclr2027_conference}

\clearpage

\appendix


{\large\textbf{Appendix}}

\section{Evaluation Details}
\label{app:evaluation}


\subsection{Benchmarks and Task Categories}

We evaluate on EB-Habitat and EB-ALFRED from EmbodiedBench~\citep{yang2025embodiedbench}, and the Search and Interact (SI) and Sequential Manipulation (SM) task families from VIGIL~\citep{chen2026done}.
These settings cover experience transfer under both high-level skill interfaces and more atomic native actions.
Table~\ref{tab:app_benchmark_overview} summarizes the evaluation scope.

\textbf{EmbodiedBench.}
EB-Habitat evaluates language-conditioned navigation and object rearrangement in Habitat 2.0~\citep{szot2021habitat}, while EB-ALFRED evaluates household tasks in AI2-THOR~\citep{kolve2017ai2}.
Each environment contains 300 episodes, with 50 episodes in each of six capability categories.
\emph{Base} evaluates basic task execution;
\emph{Common Sense} requires interpreting indirect object references through commonsense knowledge;
\emph{Complex Instruction}, corresponding to the benchmark's Complex Instruction subset, tests instruction understanding with additional context or conditional language;
\emph{Spatial Awareness} requires resolving spatial references;
\emph{Visual Appearance} identifies targets through appearance attributes;
and \emph{Long Horizon} requires longer action sequences.
We retain these category assignments throughout evaluation.

\textbf{VIGIL.}
We use 250 episodes, comprising 125 SI and 125 SM episodes.
SI requires finding an initially unseen target, approaching it, and performing the requested interaction.
SM requires an ordered manipulation sequence, such as opening a container, picking an object, and placing it at a destination.
These tasks complement EmbodiedBench by evaluating execution through more atomic native actions.

\subsection{Observation and Action Interfaces}

\textbf{EmbodiedBench.}
The agent receives the task instruction, egocentric visual observations, the available high-level action set, and textual execution feedback exposed by the benchmark.
Targets are specified through the object or receptacle identifiers supported by the action interface.
The agent selects high-level skills rather than directly controlling low-level motion.
Execution feedback indicates action validity or failure reasons and can inform subsequent decisions.

\textbf{VIGIL.}
Following its default interface, the agent receives egocentric RGB observations without action-success feedback.
It acts through task-specific native controls, including movement and pixel-grounded interaction, with a semantic report for agent-initiated termination.

\textbf{Information Access.}
Experience application does not introduce additional access to simulator information beyond the corresponding benchmark interface.
In particular, the evolving Task State $S_t$ is an agent-maintained representation derived from accessible interaction evidence, rather than a copy of the evaluator's world state.
Applicability checks and outcome verification operate on this evidence.
Evaluator-side goal predicates and success labels are used for scoring and are not supplied as online completion signals.

\subsection{Fold-Based Evaluation Protocol}

\textbf{Construction and Evaluation Splits.}
To separate the interactions used to construct experience from those used to assess transfer, we adopt a three-fold evaluation protocol.
For each environment, the episode pool is partitioned into three disjoint folds,
$\mathcal{F}_1$, $\mathcal{F}_2$, and $\mathcal{F}_3$.
In run $k$, $\mathcal{F}_k$ is held out for evaluation, while the remaining two folds provide episodes for source-trajectory collection and experience construction.
Rotating the evaluation fold ensures that each episode is evaluated once.

We construct the folds using outcome-blind grouped partitioning: predefined episode groups are kept intact, and neither episode outcomes nor method-specific performance scores are used to determine fold assignments.
For EB-Habitat and EB-ALFRED, each run uses 200 episodes for construction and 100 for evaluation.
For VIGIL, the three evaluation folds contain 84, 83, and 83 episodes, with corresponding construction pools of 166, 167, and 167 episodes, respectively.
Table~\ref{tab:app_fold_sizes} summarizes these split sizes.

For EB-Habitat, episodes sharing an identical normalized instruction or the same (scene ID, raw episode ID) pair are grouped together, yielding 267 groups.
Mixed-integer optimization assigns these groups to the three folds, with 16--17 episodes from each of the six benchmark subsets per fold.
For EB-ALFRED, episodes sharing an identical ALFRED task path or an identical normalized instruction are grouped together, yielding 130 groups.
Mixed-integer optimization assigns these groups to the three folds while balancing the distributions of the six benchmark subsets and seven ALFRED task families, as well as the number of groups across folds.

For VIGIL, all episodes sharing the same scene are assigned to the same fold.
Subject to this scene-level grouping constraint, the partition balances the SI and SM task families while minimizing imbalance in platform, difficulty, functional template, and target group.
VIGIL therefore uses scene-disjoint folds. 
For EB-Habitat and EB-ALFRED, the held-out unit remains the episode under the grouping constraints above; their evaluation is not fully scene-disjoint or task-template-disjoint.

\begin{table}[t]
\centering
\caption{
Overview of the Evaluated Environments.
Episode counts refer to the complete pools partitioned into construction and held-out evaluation folds.
}
\label{tab:app_benchmark_overview}
\scriptsize
\setlength{\tabcolsep}{5pt}
\renewcommand{\arraystretch}{1.1}
\begin{tabular}{lccc}
\toprule
\textbf{Environment}
& \textbf{Episodes}
& \textbf{Category Distribution}
& \textbf{Action Interface} \\
\midrule
EB-Habitat & 300 & $6 \times 50$ & High-level skills \\
EB-ALFRED  & 300 & $6 \times 50$ & High-level skills \\
VIGIL     & 250 & $2 \times 125$ & Native actions \\
\bottomrule
\end{tabular}
\end{table}

\begin{table}[t]
\centering
\caption{
Episode Counts under the Three-Fold Evaluation Protocol.
Construction and evaluation counts are listed in run order.
}
\label{tab:app_fold_sizes}
\scriptsize
\setlength{\tabcolsep}{5pt}
\renewcommand{\arraystretch}{1.1}
\begin{tabular}{lccc}
\toprule
\textbf{Environment}
& \textbf{Total}
& \textbf{Construction Episodes}
& \textbf{Evaluation Episodes} \\
\midrule
EB-Habitat & 300 & 200 / 200 / 200 & 100 / 100 / 100 \\
EB-ALFRED  & 300 & 200 / 200 / 200 & 100 / 100 / 100 \\
VIGIL      & 250 & 166 / 167 / 167 & 84 / 83 / 83 \\
\bottomrule
\end{tabular}
\end{table}

\textbf{Held-Out Isolation.}
Each run constructs its experience bank exclusively from source trajectories associated with its construction folds.
The held-out fold's trajectories and outcomes are excluded from that run's experience construction.
An episode may serve as a construction episode in other runs, but never in the run in which it is evaluated.
The held-out unit is the episode; this protocol does not imply scene-disjoint or task-template-disjoint evaluation.

\textbf{Frozen-Bank Evaluation.}
Once construction is complete, the experience bank is frozen for the entire evaluation run.
Evaluation interactions are not used to add, revise, or remove experience entries.
Within an episode, the agent may update $S_t$, retrieve different experiences, and revise its plan as new evidence becomes available.
These episode-local updates are reset between evaluation episodes and do not produce cross-episode experience accumulation.
Thus, the protocol evaluates transfer from a fixed source experience bank rather than continual experience acquisition during evaluation.

\textbf{Consistency across Methods.}
All compared methods use the same fold assignments and are evaluated on the same held-out episodes.

\subsection{Evaluation Metrics}

\textbf{EmbodiedBench.}
We report task success rate according to the environment's task-completion evaluator.
An episode is successful only when the complete task condition is satisfied; completing individual subgoals alone does not count as task success.
Results are reported separately for the six capability categories and for each environment overall.

\textbf{VIGIL.}
We report world-state completion $W$ rather than benchmark success $B$.
For a trajectory $\tau$ with terminal world state $s_T$ and semantic task-goal predicate $\mathcal{G}_{\mathrm{sem}}$,
\begin{equation}
W(\tau)
=
\mathbb{I}\!\left[s_T \models \mathcal{G}_{\mathrm{sem}}\right].
\label{eq:app_vigil_w}
\end{equation}
Unlike $B$, this metric does not additionally require a correct terminal report.
It measures goal satisfaction at episode closure, not whether the goal was reached at any earlier step.

\textbf{Aggregation.}
All results are reported as percentages.
We aggregate results by pooling held-out episode outcomes across the three folds, with each episode contributing exactly once.
For the pooled episode set $\mathcal{E}$ corresponding to an environment or a task category, the success rate is
\begin{equation}
\mathrm{SR}(\mathcal{E})
=
\frac{100}{|\mathcal{E}|}
\sum_{\tau \in \mathcal{E}} y_{\tau},
\label{eq:app_success_rate}
\end{equation}
where $y_{\tau}$ is the benchmark-defined task-success indicator for EmbodiedBench and $W(\tau)$ for VIGIL.
This aggregation is equivalent to averaging fold-level success rates weighted by the corresponding episode counts.
In the result tables, \emph{Avg.} denotes the arithmetic mean of the six category scores for each EmbodiedBench environment and of the SI and SM scores for VIGIL.
Since the categories within each environment have equal total episode counts, these averages coincide with the corresponding overall episode-level success rates.

Throughout evaluation, benchmark success is determined by the environment evaluator, independently of the agent's internal outcome-verification decisions.

\section{Implementation Details}
\label{app:implementation}


\subsection{Source Trajectory Collection}

For each evaluation run, source trajectories are collected from the corresponding construction folds before experience construction.
We use the same backbone model as in the corresponding benchmark evaluation and collect source trajectories under the NoExp setting, without access to any experience or memory.
Each source episode is executed once, producing a trajectory that records the task instruction, observations, actions, available execution feedback, and the final episode outcome.

Both successful and unsuccessful trajectories are used for experience construction.
Since trajectory-level success does not determine the outcomes of individual subgoal attempts, we attribute outcomes based on the local interaction evidence associated with each subgoal.
This allows verified local successes to be retained even from globally unsuccessful trajectories.

\subsection{Experience Construction}

\textbf{Subgoal Segmentation.}
The constructor $\mathcal{C}$ identifies subgoal attempts within each source trajectory and associates them with the relevant action-observation spans.
A subgoal specifies a locally assessable intermediate objective and may require one or more native actions.
Segmentation is therefore based on the attempted subgoal and its interaction evidence rather than a fixed number of actions.
Each span includes the context in which the attempt begins, the actions taken, and the resulting observations needed to assess its outcome.

\textbf{Local Outcome Attribution.}
For each attempt, the constructor determines whether the associated evidence supports the intended local outcome.
Issuing an action or following a plausible action sequence is not, by itself, sufficient evidence of completion.
Likewise, the episode-level success label cannot establish which intermediate attempts succeeded.
An attempt with insufficient evidence is not treated as a verified local success.
Conversely, a verified local success remains a valid source of action--outcome knowledge even when a later part of the source task fails.

\textbf{Experience Abstraction.}
The constructor abstracts supported subgoal attempts into experiences
$e_i=(g_i,R_i,\pi_i,P_i)$.
The abstraction separates the subgoal being pursued, the conditions under which the policy applies, the execution strategy, and the outcome that establishes local completion.
Task-specific details are retained when they constrain applicability or execution, while incidental details of the source episode are excluded from the transferable content.
The resulting experience describes a strategy to be interpreted in a new task, rather than an instruction to replay the complete source trajectory.

\textbf{Failure Utilization.}
Failure utilization operates at the subgoal level.
Verified local successes are retained independently of whether their source trajectories succeed overall.
Localized failure evidence is also preserved to inform subsequent execution, but an unsuccessful attempt is not presented as evidence that its policy achieves the intended outcome.
Similarly, a corrective strategy is supported as a successful recovery only when the source interaction contains evidence that the correction establishes the relevant local outcome.
Failure evidence alone does not establish an effective recovery policy or a definitive causal explanation of the failure.

\subsection{Experience Representation}

Table~\ref{tab:app_experience_fields} summarizes the four experience fields.
The distinction between \textsc{Requires} and \textsc{Produces} is important:
the former describes conditions that support applying the policy, whereas the latter specifies the local outcome to be assessed after execution.

\begin{table}[t]
\centering
\caption{
Fields of an Outcome-Grounded Subgoal Experience.
}
\label{tab:app_experience_fields}
\scriptsize
\setlength{\tabcolsep}{5pt}
\renewcommand{\arraystretch}{1.1}
\begin{tabular}{p{0.25\linewidth}p{0.66\linewidth}}
\toprule
\textbf{Field} & \textbf{Content} \\
\midrule
\textsc{Subgoal} ($g_i$)
& The task-relevant intermediate objective addressed by the experience. \\

\textsc{Requires} ($R_i$)
& Evidence-supported conditions under which the policy guidance is applicable. \\

\textsc{Policy} ($\pi_i$)
& Adaptable execution guidance for pursuing the subgoal, including effective behavior or condition-specific operations to avoid. \\

\textsc{Produces} ($P_i$)
& The intended local outcome that, when verified, establishes subgoal completion. \\
\bottomrule
\end{tabular}
\end{table}

\subsection{Retrieval and Experience Selection}

At each subgoal-level decision point, the executor identifies the current subgoal $\hat{g}_t$ and retrieves the top-$K$ relevant experiences from the frozen bank.
We use \texttt{all-MiniLM-L6-v2}~\citep{reimers2019sentencebert, wang2020minilm} to encode retrieval keys and queries, and rank experiences by cosine similarity in the embedding space.
For each experience, the retrieval key is its \textsc{Subgoal} field $g_i$, while the query is formed by concatenating the current task instruction $\mathcal{I}$ with the identified subgoal $\hat{g}_t$.
Unless otherwise specified, we set $K=3$.
Retrieval ranks semantic relevance only; applicability is assessed
separately using each candidate's REQUIRES field.

The retrieved top-$K$ experiences are checked in retrieval-rank order
against the current Task State $S_k$.
A candidate is applicable only when its \textsc{Requires} are supported
and its \textsc{Produces} is not already established.
The executor selects the highest-ranked applicable candidate;
if none is applicable, it proceeds without experience guidance.
These checks use agent-accessible evidence rather than the evaluator's
hidden world state.

Retrieval and selection are repeated as execution progresses.
Thus, a candidate that is not applicable at one step may become selectable later as new interaction evidence updates the Task State.

\subsection{State Maintenance, Verification, and Fallback}

\textbf{Task State.}
The evolving Task State $S_k$ summarizes task-relevant facts supported by the interaction history.
It is initialized from the current instruction and available initial observations and updated as further evidence becomes available.
Relevant facts may include object locations, possession, receptacle states, and established subgoal outcomes, subject to the information exposed by the environment.
The representation is episode-local and is reset when a new episode begins.

State maintenance distinguishes observed progress from intended effects.
Selecting an experience or issuing its actions does not automatically add $P_i$ to $S_k$.
Updates instead reflect what the resulting evidence supports, including partial progress or an unsuccessful attempt.
As execution changes the environment, previously established facts may also require revision; the current Task State is not simply the union of all outcomes observed earlier in the episode.

\textbf{Policy Application and Outcome Verification.}
The selected policy informs the executor's action choices while allowing adaptation to the current observations and available action arguments.
After execution, newly acquired interaction evidence is assessed against $P_i$.
Completion is established only when that evidence supports the intended outcome.
When the evidence is insufficient, the outcome remains unverified rather than being inferred from the fact that the policy was attempted.

The resulting state update informs subsequent planning and experience selection.
This is the basis of online composition: an established outcome may satisfy the requirements of a later experience, including one constructed from a different source trajectory.
The transition to that experience depends on the current evidence-supported state, not on a precomputed concatenation of source policies.

\textbf{Fallback and Replanning.}
When no retrieved experience is usable, the agent continues with its native planning and execution policy without conditioning on an experience.
Experience availability is therefore not a prerequisite for taking an action.
Similarly, an unsuccessful or unverified attempt does not force the agent to continue a previously selected experience sequence.
The agent updates its state from the available evidence and can replan or select a different experience as execution proceeds.

\section{Additional Experimental Results}


\begin{table}[t]
\centering
\caption{
Fold-Wise Results under the Grouped Three-Fold Evaluation Protocol.
We report the number of successful episodes on EB-Habitat and EB-ALFRED,
and the number of episodes achieving world-state completion ($W$) on VIGIL.
Overall denotes the total count across the three held-out evaluation folds.
}
\label{tab:fold_results}

\scriptsize
\setlength{\tabcolsep}{3pt}
\renewcommand{\arraystretch}{1.1}

\begin{tabular}{lcccccccccccc}
\toprule
& \multicolumn{4}{c}{\textbf{EB-Habitat}}
& \multicolumn{4}{c}{\textbf{EB-ALFRED}}
& \multicolumn{4}{c}{\textbf{VIGIL}} \\
\cmidrule(lr){2-5}
\cmidrule(lr){6-9}
\cmidrule(lr){10-13}

\textbf{Method}
& Fold 1 & Fold 2 & Fold 3 & \textbf{Overall}
& Fold 1 & Fold 2 & Fold 3 & \textbf{Overall}
& Fold 1 & Fold 2 & Fold 3 & \textbf{Overall} \\
\midrule

\textit{Eval. episodes}
& 100 & 100 & 100 & 300
& 100 & 100 & 100 & 300
& 84  & 83  & 83  & 250 \\
\midrule

NoExp
& 44 & 55 & 51 & 150
& 54 & 60 & 50 & 164
& 23 & 30 & 26 & 79 \\

A-Mem
& 52 & 61 & 52 & 165
& 58 & 62 & 58 & 178
& 24 & 30 & 34 & 88 \\

Mem0
& 49 & 62 & 53 & 164
& 58 & 64 & 54 & 176
& 26 & 27 & 28 & 81 \\

LangMem
& 54 & 62 & 58 & 174
& 58 & 52 & 55 & 165
& 27 & 29 & 30 & 86 \\

ReMe
& 48 & 57 & 56 & 161
& 55 & 62 & 55 & 172
& 24 & 34 & 33 & 91 \\

WorldMind
& 44 & 56 & 55 & 155
& 55 & 57 & 55 & 167
& 25 & 30 & 31 & 86 \\

EmbodiSkill
& 44 & 59 & 54 & 157
& 52 & 46 & 55 & 153
& 28 & 25 & 31 & 84 \\

\textbf{EmbodiExp (Ours)}
& \textbf{65} & \textbf{71} & \textbf{68} & \textbf{204}
& \textbf{63} & \textbf{69} & \textbf{73} & \textbf{205}
& \textbf{29} & \textbf{36} & \textbf{37} & \textbf{102} \\

\bottomrule
\end{tabular}
\end{table}

\subsection{Detailed Quantitative Results}

Table~\ref{tab:fold_results} provides fold-wise results for all methods using the same configurations as the main experiments.
Each episode is evaluated exactly once, with experience banks constructed from the other two folds and frozen during evaluation.
Performance remains broadly comparable across folds, with \nickname outperforming all baselines in every fold on all three benchmarks.
This consistency suggests that the overall gains are not driven by a particularly favorable split, supporting the stability of the comparisons under the grouped three-fold evaluation protocol.

\subsection{Efficiency Analysis}

\begin{table}[t]
\centering
\caption{
Online Efficiency on EB-Habitat with Qwen3.6-35B-A3B.
Except for SR, all metrics are per-episode averages over
300 held-out episodes, including failures.
Model calls include all online logical invocations, retries,
and cache hits.
Tokens sum the corresponding server-reported input and output
usage, using saved usage for cached responses.
Offline experience construction is excluded.
}
\label{tab:efficiency}

\scriptsize
\setlength{\tabcolsep}{8pt}
\renewcommand{\arraystretch}{1.1}

\begin{tabular}{lcccc}
\toprule
\textbf{Method}
& \textbf{SR (\%)} $\uparrow$
& \textbf{Env. steps}
& \textbf{Model calls}
& \textbf{Tokens (k)} \\
\midrule
NoExp
& 50.0 & 15.96 & 7.34 & 37.84 \\
\textbf{\nickname (Ours)}
& \textbf{68.0} & 13.67 & 17.44 & 77.22 \\
\bottomrule
\end{tabular}
\end{table}

Table~\ref{tab:efficiency} examines the balance between online model usage and environment interaction. 
Compared with NoExp, \nickname improves task success from 50.0\% to 68.0\% while reducing average environment steps from 15.96 to 13.67.
These improvements are accompanied by approximately $2.38\times$ as many logical model calls and $2.04\times$ as many reported tokens, including the additional calls for experience selection and local outcome verification.
Overall, \nickname balances increased model usage with substantially higher task completion and fewer environment interactions, highlighting the trade-off between model-side reasoning and environment-side execution.

\subsection{Additional Qualitative Results}

\begin{figure*}[t]
\centering
\centerline{
\includegraphics[width=\textwidth]{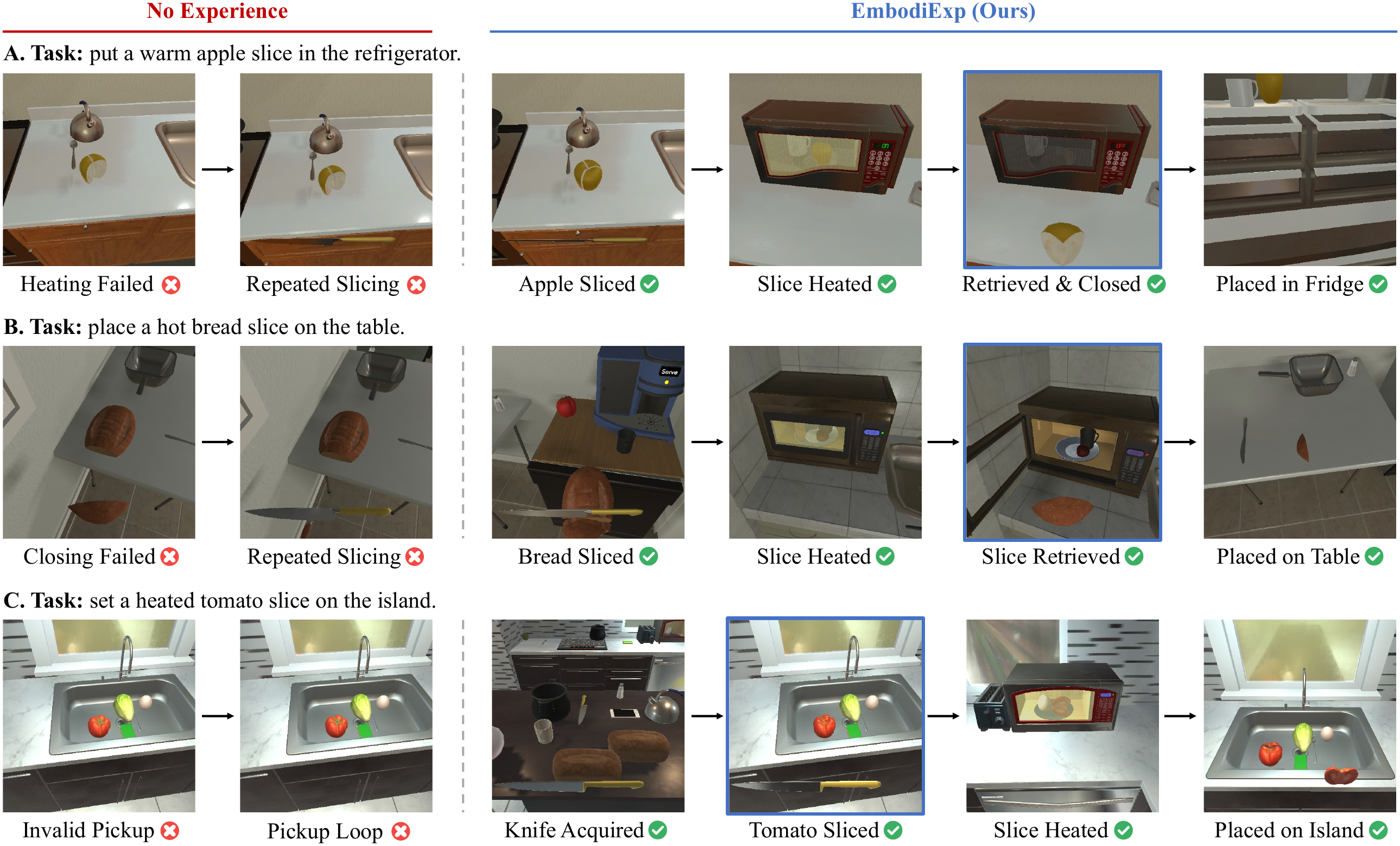}
}
\caption{
Qualitative Results on EB-ALFRED.
In A and B, NoExp returns to repeated slicing after microwave-operation failures,
whereas EmbodiExp retrieves the heated food and completes placement.
In C, NoExp repeats invalid pickups, whereas EmbodiExp proceeds through slicing,
heating, and placement.
Blue outlines mark verified local outcomes during experience application.
}
\label{fig:qualitative_alfred}
\vspace{-0.3cm}
\end{figure*}

Figure~\ref{fig:qualitative_alfred} presents three paired long-horizon manipulation examples on EB-ALFRED.
In \ref{fig:qualitative_alfred}.A and \ref{fig:qualitative_alfred}.B, NoExp returns to repeated slicing after microwave-operation failures, whereas \nickname retrieves the heated food with verified local outcomes and completes placement.
In \ref{fig:qualitative_alfred}.C, NoExp repeats invalid pickups, while \nickname acquires the knife, completes slicing with a verified outcome, and proceeds to heating and placement.
These examples illustrate how local experience supports intermediate state transitions, with verified outcomes providing a basis for continued progress across subgoals.

\begin{figure*}[t]
\centering
\centerline{
\includegraphics[width=\textwidth]{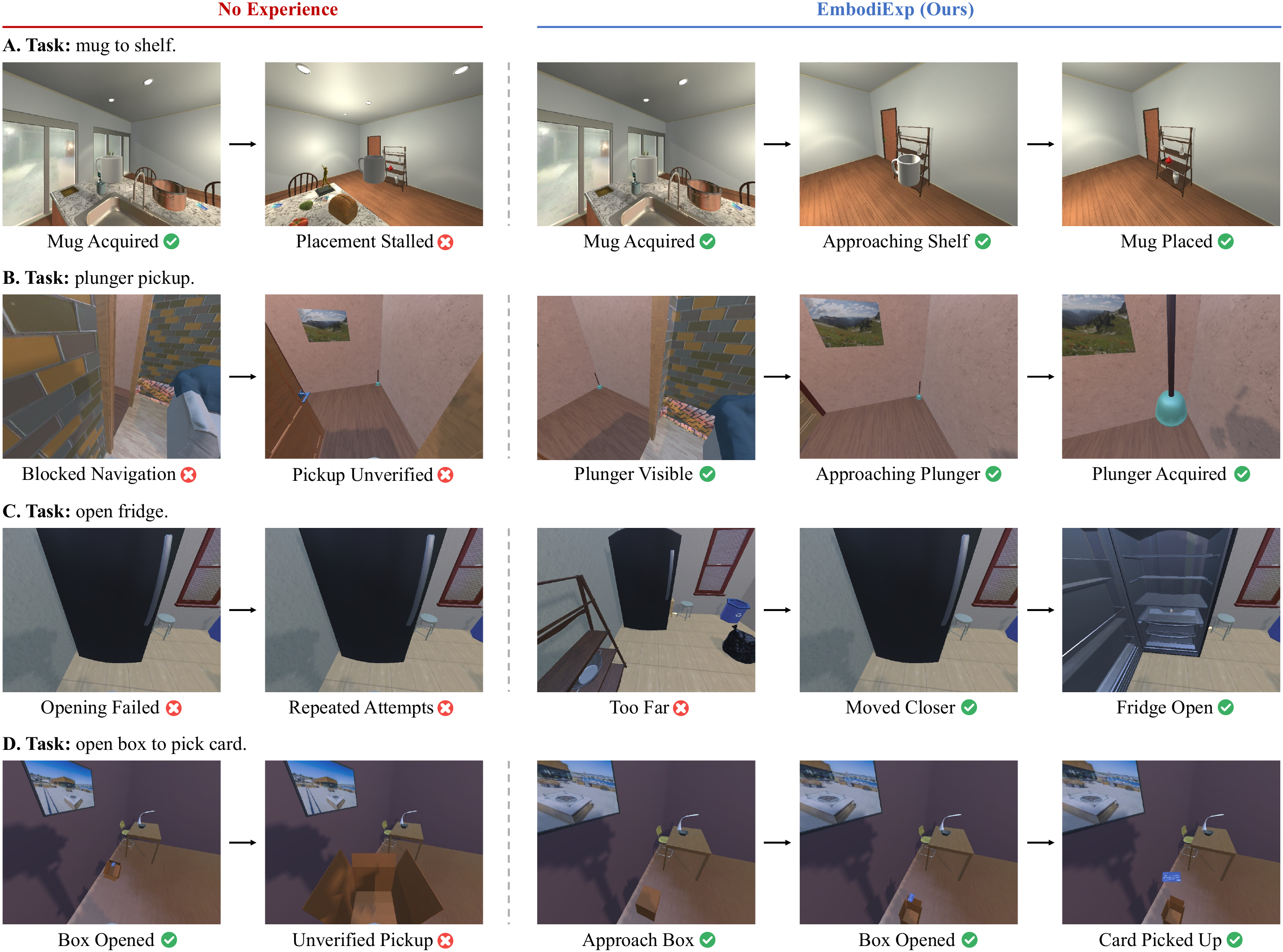}
}
\caption{
Qualitative Results on VIGIL.
In A and B, \nickname progresses from pickup to placement and from search to pickup, respectively, while NoExp stalls during navigation or interaction.
In C, NoExp repeats unsuccessful opening attempts, whereas \nickname recovers by moving closer before opening the fridge.
In D, both agents open the box, but only \nickname completes the subsequent card pickup.
}
\label{fig:qualitative_vigil}
\vspace{-0.3cm}
\end{figure*}

Figure~\ref{fig:qualitative_vigil} presents four paired examples on VIGIL.
In \ref{fig:qualitative_vigil}.A and \ref{fig:qualitative_vigil}.B, \nickname connects verified object possession and target visibility to subsequent placement and pickup, respectively, whereas NoExp stalls during navigation or interaction.
In \ref{fig:qualitative_vigil}.C, it moves within reach before retrying the failed opening action, rather than repeating ineffective attempts.
In \ref{fig:qualitative_vigil}.D, it verifies the open-container prerequisite and completes the card pickup, while NoExp reports completion after an unverified pickup.
These selected cases illustrate how local experience supports subgoal transitions and interaction recovery, with achieved intermediate states providing a concrete basis for subsequent actions.

\section{Limitations and Future Work}
\label{app:limitations}

\textbf{Frozen-Bank Evaluation.}
We intentionally freeze the experience bank during evaluation to focus on what experience to retain and how to apply it across tasks, separating these questions from continual experience acquisition.
While experience selection and composition remain online, the source bank is fixed, and the current evaluation does not examine its evolution through subsequent interactions.
We leave extending \nickname to online self-evolving embodied agents that continually acquire, refine, and apply experience as they encounter new tasks for future work.

\textbf{Simulation and Task-Level Execution.}
Our experiments focus on task-level embodied execution in simulated environments.
Although the evaluation covers both high-level skills and more atomic native actions, it does not yet examine experience transfer within vision-language-action (VLA) policies for low-level robot control.
Future work includes extending the evaluation to VLA benchmarks with finer-grained control interfaces, with real-world robotic interaction as a further direction beyond the current simulation-based setting.

\textbf{Scale and Diversity of Source Experience.}
Our experience banks are constructed from benchmark-specific source trajectory pools under a controlled collection protocol.
This setting supports held-out evaluation with a consistent source pool, but does not characterize experience transfer from substantially larger and more heterogeneous collections of interactions.
Future work could expand the scale and diversity of source trajectories, including interactions collected by different agents and policies, and investigate how these factors affect experience construction and transfer to new tasks.

\section{Prompt Examples}
\label{app:prompts}

\begingroup
\setlength{\emergencystretch}{2em}

\newcounter{AppPromptNumber}
\newbox\AppPromptBody
\newbox\AppPromptPart
\newbox\AppPromptHead
\newbox\AppPromptPanel
\newdimen\AppPromptWidth
\newdimen\AppPromptOuter
\newdimen\AppPromptRoom
\newdimen\AppPromptHeight
\newdimen\AppPromptBand
\newdimen\AppPromptDim
\newif\ifAppPromptContinued

\newcommand{\AppPromptLiteral}[1]{%
  \ifdefined\pdfliteral
    \pdfliteral{#1}%
  \else\ifdefined\pdfextension
    \pdfextension literal{#1}%
  \else
  \fi\fi
}
\newcommand{\AppPromptNumberDim}[2]{%
  \AppPromptDim=\dimexpr#2\relax
  \edef#1{\csname strip@pt\endcsname\AppPromptDim}%
}
\newcommand{\AppPromptRoundPath}{%
  5 0 m \AppWR\space 0 l
  \AppWK\space 0 \AppW\space 2.2386 \AppW\space 5 c
  \AppW\space \AppHR\space l
  \AppW\space \AppHK\space \AppWK\space \AppH\space \AppWR\space \AppH\space c
  5 \AppH\space l
  2.2386 \AppH\space 0 \AppHK\space 0 \AppHR\space c
  0 5 l 0 2.2386 2.2386 0 5 0 c h
}
\newcommand{\AppPromptPaint}{%
  \AppPromptNumberDim\AppW{\AppPromptOuter}%
  \AppPromptNumberDim\AppH{\ht\AppPromptPanel+\dp\AppPromptPanel}%
  \AppPromptNumberDim\AppWR{\AppPromptOuter-5pt}%
  \AppPromptNumberDim\AppWK{\AppPromptOuter-2.2386pt}%
  \AppPromptNumberDim\AppHR{\ht\AppPromptPanel+\dp\AppPromptPanel-5pt}%
  \AppPromptNumberDim\AppHK{\ht\AppPromptPanel+\dp\AppPromptPanel-2.2386pt}%
  \AppPromptNumberDim\AppBand{\AppPromptBand}%
  \AppPromptNumberDim\AppBandY{\ht\AppPromptPanel+\dp\AppPromptPanel-\AppPromptBand}%
  \AppPromptLiteral{%
    q 0.996264 0 0 0.996264 0 0 cm
      q \AppPromptRoundPath\space W n
        0.985 0.991 1 rg 0 0 \AppW\space \AppH\space re f
        0.34 0.49 0.66 rg 0 \AppBandY\space \AppW\space \AppBand\space re f
      Q
      0.34 0.49 0.66 RG 0.7 w \AppPromptRoundPath\space S
    Q}%
}
\newcommand{\AppPromptMeasureRoom}{%
  \AppPromptRoom=\pagegoal
  \advance\AppPromptRoom by -\pagetotal
  \ifdim\AppPromptRoom>\textheight\AppPromptRoom=\textheight\fi
  \advance\AppPromptRoom by -\AppPromptBand
  \advance\AppPromptRoom by -21pt
}
\newcommand{\AppPromptRender}{%
  \par\penalty0
  \setbox\AppPromptHead=\vbox{%
    \begingroup
    \hsize=\AppPromptWidth
    \normalfont\normalsize
    \parindent=0pt\parskip=0pt\rightskip=0pt plus1fil
    \ifdefined\color\color{white}\else\AppPromptLiteral{1 1 1 rg}\fi
    \noindent\strut Prompt~\theAppPromptNumber:\enspace\AppPromptTitle
    \ifAppPromptContinued\ \textnormal{(continued)}\fi
    \strut\par
    \ifdefined\color\else\AppPromptLiteral{0 0 0 rg}\fi
    \endgroup
  }%
  \AppPromptBand=\dimexpr\ht\AppPromptHead+\dp\AppPromptHead+12pt\relax
  \AppPromptMeasureRoom
  \AppPromptHeight=\dimexpr\ht\AppPromptBody+\dp\AppPromptBody\relax
  \AppPromptDim=\dimexpr\textheight-\AppPromptBand-21pt\relax
  \ifdim\AppPromptHeight>\AppPromptRoom
    \ifdim\AppPromptHeight<\AppPromptDim
      \newpage\AppPromptMeasureRoom
    \else
      \ifdim\AppPromptRoom<120pt\newpage\AppPromptMeasureRoom\fi
    \fi
  \fi
  \ifAppPromptContinued\else
    \addtocounter{AppPromptNumber}{-1}%
    \refstepcounter{AppPromptNumber}%
    \expandafter\label\expandafter{\AppPromptLabel}%
  \fi
  \ifdim\AppPromptHeight>\AppPromptRoom
    \splittopskip=\topskip\splitmaxdepth=4pt\vbadness=10000
    \setbox\AppPromptPart=\vsplit\AppPromptBody to\AppPromptRoom
    \setbox\AppPromptPart=\vbox{\unvbox\AppPromptPart}%
  \else
    \setbox\AppPromptPart=\box\AppPromptBody
  \fi
  \setbox\AppPromptPanel=\vbox{%
    \hsize=\AppPromptOuter\offinterlineskip
    \vskip6pt
    \hbox{\kern11pt\box\AppPromptHead\kern11pt}%
    \vskip14pt
    \hbox{\kern11pt\box\AppPromptPart\kern11pt}%
    \vskip10pt
  }%
  \noindent\hbox{\rlap{\AppPromptPaint}\box\AppPromptPanel}\par
  \ifvoid\AppPromptBody\else
    \newpage\AppPromptContinuedtrue
    \expandafter\AppPromptRender
  \fi
}
\newenvironment{AppPromptText}[2]{%
  \par\addvspace{10pt}%
  \stepcounter{AppPromptNumber}%
  \def\AppPromptLabel{#1}\def\AppPromptTitle{#2}%
  \AppPromptContinuedfalse
  \AppPromptOuter=\linewidth
  \AppPromptWidth=\dimexpr\AppPromptOuter-22pt\relax
  \setbox\AppPromptBody=\vbox\bgroup
    \hsize=\AppPromptWidth\linewidth=\AppPromptWidth
    \normalfont\normalsize
    \parindent=0pt\parskip=5pt
    \leftskip=0pt\rightskip=0pt\parfillskip=0pt plus1fil
    \emergencystretch=2em
    \clubpenalty=10000\widowpenalty=10000
    \noindent\ignorespaces
}{%
    \par\unskip
  \egroup
  \AppPromptRender
  \par\addvspace{12pt}%
}
\newcommand{\AppInput}[1]{\ensuremath{\langle}\emph{#1}\ensuremath{\rangle}}

We provide prompt examples for experience construction and experience-guided execution below.
Angle brackets denote dynamic inputs adapted to each benchmark and environment.

\subsection{Offline Experience Construction}
\label{app:prompts_construction}

The experience constructor segments source trajectories into local attempts, extracts candidate experiences, and reviews them for evidence-based admission.
Each attempt may contribute zero or more entries under the shared four-field representation.

\begin{AppPromptText}{prompt:segmentation}{Trajectory Segmentation}
Segment the trajectory into coherent attempts at local objectives using the task, actions, feedback, and observations.

Keep continuous pursuit of the same objective together, including search, retries, and alternative candidates. Split when the objective changes, and do not merge attempts separated by work on another objective. Acquisition and delivery should be treated as distinct objectives when they have separate outcomes. Failure or movement alone does not define a boundary.

Include both successful and unsuccessful attempts. Cover all recorded actions with ordered, nonoverlapping, contiguous spans using zero-based inclusive step indices. Propose only the subgoal and its boundaries; do not label success or explain outcomes.

Write each subgoal as a self-contained description of the intended local change and its target. Preserve attributes, destination relations, or task constraints needed to distinguish the objective. Clearly distinguish outcomes such as reaching a location, acquiring an object, and delivering it. Omit unrelated task clauses, incidental locations, detailed preconditions, and action routes. Use concise and consistent wording without inventing details.

Return JSON with \texttt{attempts}, where each attempt contains \texttt{goal}, \texttt{start\_step}, and \texttt{end\_step}.

\AppInput{task, available\_actions, and public\_steps}
\end{AppPromptText}

Experience extraction is applied to each segmented attempt.
Demonstrations are drawn only from the construction folds, excluding the held-out fold and current source trajectory.

\begin{AppPromptText}{prompt:extraction}{Outcome-Grounded Experience Extraction}
Extract transferable experience entries from the proposed local attempt.
Use the task, actions, feedback, and observations as evidence; treat the proposed goal and boundaries as candidates to verify. Apply the same local evidence rules regardless of trajectory-level success or failure.

\textbf{1. Grounding:} identify the intended local objective and distinguish it from attempted actions and observed outcomes. Verified local completion supports guidance on effective behavior. Explicit evidence of failed operations supports condition-specific guidance on operations to avoid. Unverified completion alone does not establish failure. A completed attempt may support both forms of guidance; assess each retained claim against the relevant operations rather than the attempt's final outcome alone.

\textbf{2. Coherence:} represent each entry with \texttt{subgoal}, \texttt{requires}, \texttt{policy}, and \texttt{produces}. \texttt{Requires} specifies the supported conditions under which the guidance applies. \texttt{Policy} describes effective behavior or condition-specific operations to avoid. \texttt{Produces} specifies the intended local completion condition, not necessarily an outcome achieved in the source attempt. Do not replace the intended outcome with an observed failure or a weaker objective.

For effective behavior, ground \texttt{requires} in conditions preceding the retained behavior; do not include conditions that the policy itself establishes. For avoidance guidance, ground \texttt{requires} in the observed conditions under which the operation failed, including relevant in-span context. Do not infer a successful alternative or an unsupported cause of failure.

\textbf{3. Transferability:} retain useful decisions and guidance while removing incidental source-specific details. Preserve conditions and task constraints needed for the guidance to remain valid.

Return only JSON with an \texttt{experiences} list. Each entry contains exactly \texttt{subgoal}, \texttt{requires}, \texttt{policy}, and \texttt{produces}. Write \texttt{subgoal} as a concise, self-contained objective, \texttt{requires} and \texttt{produces} as lists of conditions, and \texttt{policy} as 1--3 concise pieces of guidance. An attempt may yield multiple entries with different applicability conditions. Return \texttt{\{"experiences": []\}} when neither effective behavior nor avoidance guidance is supported.

\AppInput{fold-filtered demonstrations}

Extract the experiences for the input below.

\begingroup
\parskip=0pt
\{\par
\hspace*{1em}``task": \AppInput{task},\par
\hspace*{1em}``proposed\_attempt": \AppInput{segmented attempt},\par
\hspace*{1em}``public\_span": \AppInput{actions, observations, and feedback within START..END},\par
\hspace*{1em}``preceding\_context": \AppInput{public interaction evidence before START},\par
\hspace*{1em}``available\_actions": \AppInput{native action names},\par
\hspace*{1em}``entry\_public\_facts": \AppInput{facts supported before START},\par
\hspace*{1em}``end\_public\_facts": \AppInput{facts supported at END},\par
\hspace*{1em}``START": \AppInput{start step},\par
\hspace*{1em}``END": \AppInput{end step}\par
\}
\endgroup
\end{AppPromptText}

Each extracted candidate is reviewed separately.
Only admitted four-field entries are added to the experience bank.

\begin{AppPromptText}{prompt:admission}{Evidence-Based Admission Review}
Review one candidate experience against the supplied public evidence. Do not rewrite the candidate or infer unsupported claims from trajectory-level success or failure.

Assess whether the proposed local objective is relevant to the task (\texttt{task\_relation}) and whether the evidence verifies its completion within START..END (\texttt{local\_outcome}). Use \texttt{segmentation\_issue} when the proposed goal and span are incoherent. An unverified local outcome does not by itself imply a failed operation or require rejection of avoidance guidance.

Copy the candidate into \texttt{checked\_claims} and evaluate all four fields regardless of \texttt{local\_outcome}:

\textbf{subgoal:} whether it faithfully represents the task-relevant local objective, without replacing it with a weaker achieved objective.

\textbf{requires:} whether the conditions are supported and appropriate for applying the guidance. For effective behavior, use evidence preceding the retained behavior and exclude conditions established by the policy itself. For avoidance guidance, verify the observed conditions preceding the failed operation; relevant evidence may occur within the attempt. Report necessary \texttt{missing\_conditions}.

\textbf{policy:} whether each piece of guidance is grounded in the relevant local operations. Effective behavior requires evidence that the retained behavior established \texttt{produces}, with necessary operations preserved; report any \texttt{missing\_operations}. Avoidance guidance requires explicit evidence of the operation's failure under \texttt{requires}, but does not require a demonstrated successful alternative. A completed attempt does not validate every operation within it. Reject unsupported success claims, alternatives, or causal explanations.

\textbf{produces:} whether it specifies the intended, locally verifiable completion condition for \texttt{subgoal}. Do not require it to have been achieved in the source attempt when the entry provides only evidence-supported avoidance guidance.

Return JSON with \texttt{task\_relation}, \texttt{local\_outcome}, \texttt{segmentation\_issue}, \texttt{checked\_claims}, \texttt{experience\_checks}, and a Boolean \texttt{admit}. For each evidence assessment, report a \texttt{status} (\texttt{supported}, \texttt{contradicted}, or \texttt{unknown}), a \texttt{reason}, and \texttt{evidence\_steps}. Evaluate \texttt{experience\_checks} even when completion is unverified. Admit only when the objective, applicability conditions, guidance, and intended completion condition pass their respective checks; reject entries with unsupported guidance or unresolved critical checks. Assess factual and operational correctness rather than stylistic quality. Review metadata is not part of the stored four-field experience.

\AppInput{task, proposed attempt, public evidence, and candidate experience}
\end{AppPromptText}

\subsection{Online Experience-Guided Execution}
\label{app:prompts_execution}

At each planning step, the agent maintains the Task State, identifies the current subgoal, and selects the highest-ranked applicable experience.
Its policy guides native action generation.
After the resulting execution span, local outcome verification and newly acquired evidence inform the next Task State update.

\begin{AppPromptText}{prompt:state}{Task State and Current-Subgoal Identification}
Update the public Task State from the available interaction evidence and identify one useful next local subgoal toward completing the human task. At episode start, initialize the state from the instruction and initial observation. Express the subgoal as a meaningful local outcome, leaving native navigation and interaction commands to the actor.

Distinguish desired task conditions from observed facts. Ground facts in observations and available executed-action feedback while preserving task constraints and object/destination distinctions. Carry previous facts forward only while they remain supported; revise or remove facts invalidated by newer evidence.

Credit an experience's intended outcome only when verification supports it. When completion remains unverified, still incorporate other supported changes, including partial progress. Selecting an experience or issuing an action does not establish its intended outcome. Keep uncertainty explicit, do not equate unverified completion with failure, and do not infer whole-task completion from local progress.

Use the supplied robot appearance information to distinguish robot hardware from manipulated objects. Visible arm or gripper components do not imply that an object is held. Determine holding status from visual evidence and available pickup or release feedback; report uncertainty when ambiguous.

Return JSON with \texttt{reason}, \texttt{facts}, \texttt{current\_subgoal}, and \texttt{uncertainties}. Include at most eight concise facts, each with \texttt{claim} and \texttt{evidence\_steps}. Use public step indices for prior evidence and \texttt{-1} for the current RGB observation. Keep the state episode-local; do not update the frozen experience bank.

\AppInput{task, current RGB observation, available\_actions, public\_steps, explicit\_feedback\_facts, previous\_facts, local verification results when available, verified\_local\_events, and robot appearance information}
\end{AppPromptText}

\begin{AppPromptText}{prompt:binding}{Applicability Checking and Task-Specific Binding}
Select at most one experience for the current subgoal from the retrieved candidates, preserving their supplied retrieval order.

For each candidate, bind \texttt{subgoal}, \texttt{requires}, \texttt{policy}, and \texttt{produces} to the current task where justified by the task and current evidence. Preserve the transferable guidance and necessary conditions. Do not invent entities, change the strategy, or weaken task constraints to make a candidate applicable. Binding creates an episode-local application and does not modify the stored experience.

Check each bound requirement against the current Task State. Mark it \texttt{supported}, \texttt{contradicted}, or \texttt{unknown}. Expected policy effects do not count as starting evidence. A candidate is applicable only when every requirement is supported and its intended \texttt{produces} is not already established. \texttt{Produces} is established only when all of its completion conditions are supported. Reject candidates with any contradicted or unknown requirement, or with an already established intended outcome.

Select the highest-ranked applicable candidate. Do not rerank applicable candidates by an additional preference judgment. If none is applicable, return no experience so that the agent can act without experience guidance.

Return JSON with \texttt{reason}, \texttt{selected\_id}, \texttt{application}, \texttt{requires\_checks}, and \texttt{useful\_unresolved\_work}. Each requirement check contains \texttt{claim}, \texttt{reason}, \texttt{status}, and \texttt{evidence\_steps}. The selected \texttt{application} contains the four bound experience fields; keep its \texttt{subgoal} and \texttt{produces} fixed during the ensuing execution and verification. If no experience is selected, set \texttt{selected\_id} and \texttt{application} to null and explain the rejection reasons briefly.

\AppInput{task, current subgoal, current Task State, available actions, public steps, and top-3 retrieved experiences in retrieval-rank order}
\end{AppPromptText}

The native actor uses its system prompt with either the initial-planning or replanning suffix.
The current Task State and subgoal are supplied for planning; the experience-guidance suffix is appended only when an experience is selected.
When no candidate is applicable, the actor proceeds without experience guidance.
Native action definitions and task-execution examples are supplied through the corresponding inputs.

\begin{AppPromptText}{prompt:actor-system}{Native Actor: System Prompt}
You are a robot operating in an environment. Complete the given task using only the defined actions.

\textbf{Action Descriptions and Validity Rules}

\AppInput{native action descriptions and validity rules}

\textbf{Available Actions}

\AppInput{available actions}

\AppInput{in-context task-execution examples}

\textbf{Guidelines}

Produce a nonempty plan of at most 20 valid actions. Match each action name with its corresponding action ID and respect all action validity rules.

Use navigation or search actions when required objects or receptacles are not visible. When multiple instances of an object exist, explore alternatives as needed.

Use the interaction history and environment feedback to adapt the plan. If an action fails or is invalid, revise the strategy rather than repeating the same unsuccessful action sequence.
\end{AppPromptText}

\begin{AppPromptText}{prompt:actor-initial}{Native Actor: Initial Planning Suffix}
The human instruction is: \AppInput{task instruction}.

Return JSON describing the current visual state, reasoning and reflection, a language-level plan, and the executable action sequence.

Use the fields \texttt{visual\_state\_description}, \texttt{reasoning\_and\_reflection}, \texttt{language\_plan}, and \texttt{executable\_plan}. Each executable action contains its \texttt{action\_id} and exact \texttt{action\_name}.
\end{AppPromptText}

\begin{AppPromptText}{prompt:actor-replan}{Native Actor: Replanning Suffix}
The human instruction is: \AppInput{task instruction}.

The action history is:
\AppInput{step-indexed action IDs, action names, and environment feedback}

Based on the interaction history, current Task State, environment feedback, and current visual state, revise the plan from the current state. Continue from supported progress and address unresolved objectives. When evidence shows that an operation failed or was invalid, revise the relevant behavior rather than repeating it unchanged. Do not assume failure merely because completion remains unverified, and do not invent a cause unsupported by the evidence.

Return JSON with \texttt{visual\_state\_description}, \texttt{reasoning\_and\_reflection}, \texttt{language\_plan}, and \texttt{executable\_plan}. Each executable action contains its \texttt{action\_id} and exact \texttt{action\_name}.
\end{AppPromptText}

\begin{AppPromptText}{prompt:actor-guidance}{Native Actor: Appended Experience Guidance}
The selected experience provides the following policy for the current local objective:

\AppInput{task-bound experience policy}

Use this policy as supplementary guidance when generating the next actions. It may describe effective behavior or condition-specific operations to avoid. Apply avoidance guidance only under its stated conditions; it does not establish that an untested alternative will succeed.

Adapt action choices and arguments to the current observations, available actions, and environment feedback. Preserve all constraints of the human task, and do not follow guidance that conflicts with current evidence or action validity rules. Following a policy does not establish completion; the intended outcome must be verified from the resulting interaction evidence.
\end{AppPromptText}

The following prompt verifies the selected experience after the execution span at a planning step, before the subsequent Task State update.
Verification credits the intended outcome only when supported; other supported facts may update the state regardless of the verification result, while invalidated facts are revised or removed.

\begin{AppPromptText}{prompt:verification}{Local Outcome Verification}
Verify the frozen local experience application after the current execution span, which may contain one or more native actions. Use only agent-accessible observations and actual environment feedback collected over the span; do not use hidden evaluator state or task-success labels.

Check the local subgoal and each \texttt{produces} claim in the supplied order. Mark each claim \texttt{supported}, \texttt{contradicted}, or \texttt{unknown}, accounting for later evidence that invalidates earlier observations. An attempted action, a plausible plan, or an expected effect is not evidence of completion. Missing support alone is not a contradiction.

Preserve the original identities, destinations, and task constraints. Do not reinterpret or weaken the frozen application to declare success. Assess local completion independently of whole-task success. The intended outcome is verified only when the evidence supports \texttt{produces} in full; otherwise completion remains unverified, which does not by itself establish failure.

Return JSON with an ordered \texttt{checks} list. Each check contains the unchanged \texttt{claim}, \texttt{reason}, \texttt{status}, and \texttt{evidence\_steps}. Use public step indices for recorded evidence and \texttt{-1} for the current RGB observation. Report support for individual claims even when the complete intended outcome remains unverified.

\AppInput{task, frozen application, ordered subgoal-and-produces claims, application start step, public evidence from the execution span, current RGB observation, and explicit feedback facts}
\end{AppPromptText}

\endgroup

\end{document}